\documentclass{article} % For LaTeX2e
\usepackage{preprint,times}

\usepackage{amsmath,amsfonts,bm}

\def\eqref#1{equation~\ref{#1}}
\def\1{\bm{1}}

\DeclareMathAlphabet{\mathsfit}{\encodingdefault}{\sfdefault}{m}{sl}
\SetMathAlphabet{\mathsfit}{bold}{\encodingdefault}{\sfdefault}{bx}{n}

\usepackage{hyperref}
\usepackage{url}

\usepackage[T1]{fontenc}
\usepackage{booktabs}
\usepackage{multirow}
\usepackage{enumitem}
\usepackage{graphicx}
\usepackage{subcaption}
\usepackage{amsmath}
\usepackage{mdframed}
\usepackage{soul}
\usepackage{ulem}
\usepackage{tcolorbox}
\tcbuselibrary{skins, theorems}

\usepackage{booktabs, multirow}
\usepackage{tcolorbox}
\usepackage{adjustbox}
\usepackage{amsthm}

\definecolor{raw}{HTML}{9A9A9A}
\definecolor{grounded}{HTML}{1E88E5}
\definecolor{groundedglobal}{HTML}{F9A40C}
\definecolor{degraded}{HTML}{D81B60}

\theoremstyle{definition}
\newtheorem{definition}{Definition}

\tcbuselibrary{breakable}

\newtcolorbox{promptbox}[1][]{
  breakable,
  enhanced,
  colback=gray!5,
  colframe=gray!50,
  title={#1},
  fontupper=\ttfamily\footnotesize,
}

\definecolor{defbg}{HTML}{FEF3ED}
\definecolor{defframe}{HTML}{C58F86}
\definecolor{deftitle}{HTML}{5C2F21}

\definecolor{defbgB}{HTML}{EDFEEF}
\definecolor{defframeB}{HTML}{86C58C}
\definecolor{deftitleB}{HTML}{215C26}

\newcommand{\dataset}{\textsc{MIMIC-III-Ext-BHC-Edit}}
\newcommand{\verifactbhc}{\textsc{VeriFact-BHC}}

\usetikzlibrary{positioning}

\newif\ifshowcomments
\showcommentstrue

\newif\ifshowst
\showstfalse

\title{Actionable Feedback: Improving Model\\Capabilities from User Edit Data on Clinical\\Documentation}

\author{Daniel P. Jeong$^{\ast\dagger}$ \\ Carnegie Mellon University\\Abridge AI
\And Yewon Byun$^{\ast\dagger}$ \\ Carnegie Mellon University\\Abridge AI
\And John Giorgi \\ Abridge AI
\AND Peng Xu \\ Abridge AI
\And Zachary C. Lipton \\ Carnegie Mellon University
\And Michael Oberst \\ Johns Hopkins University\\Abridge AI
}

\iclrfinalcopy
\begin{document}

\maketitle
\renewcommand{\thefootnote}{\fnsymbol{footnote}}
\footnotetext[1]{Equal contribution.}
\footnotetext[2]{Work conducted as interns at Abridge AI.}
\renewcommand{\thefootnote}{\arabic{footnote}}

\begin{abstract}
\vspace{-7pt}
Large language models (LLMs) are increasingly being deployed in clinical documentation workflows, where clinicians review and edit their outputs before use. In principle, these clinician edits represent a scalable source of supervision for model training. In practice, not all edits represent \textit{actionable} feedback: for instance, on data drawn from real-world deployment of an AI clinical documentation product, our review suggests that over 45\% of edited outputs contain user-added information that is not directly supported by the inputs to the model. Naively fine-tuning on such edits risks teaching the model to hallucinate. In this paper, we develop a method to address this challenge. Given that real clinician edits on deployed documentation tools are not publicly available, we first develop a taxonomy of clinician-editing behaviors, informed by data from a deployed AI clinical documentation product. Then, we develop a method for generating semi-synthetic clinician-edit data, informed by our taxonomy, on a discharge-summary generation task based on MIMIC-III, a public dataset. Finally, we introduce simple filtering methods for identifying \textit{actionable} edits, with accuracy evaluated by expert review. We show empirically, on our semi-synthetic dataset, that naive fine-tuning on raw edited output \textit{increases} ungrounded content, while our approach of progressively filtering toward grounded edits enables improvement on clinician-alignment metrics while retaining faithfulness to the input context. 
\end{abstract}

\section{Introduction}
\label{sec:intro}
\vspace{-7pt}
A central challenge in clinical documentation is summarizing large volumes of patient information into notes that clinicians can readily use for patient care. 
Large language model (LLM)-based applications are increasingly used to automate this process, 
ranging from ambient scribes that draft patient-visit summaries from clinician-patient conversations~\citep{tierney2024-10,lukac2025-76,afshar2025-66} to systems that generate discharge summaries from longitudinal inpatient records~\citep{llm-physician-discharge-summary,safety-outcomes-bhc} or draft patient portal messages~\citep{physician-edits-messages}. 
The resulting LLM outputs are closely reviewed and revised by a clinician in the loop before being finalized as documents in the electronic health record (EHR) or as messages to send to patients. 
In contrast to other summarization or generation tasks, where user feedback may be limited to coarse ratings of quality,
these settings provide a rich environment for learning: 
expert edits are routinely collected during deployment and reveal the gaps between what the model produced and what a clinician would accept, 
providing, in theory, ideal supervision for 
training models. 

\begin{figure}[t]
    \centering
    \includegraphics[width=\linewidth]{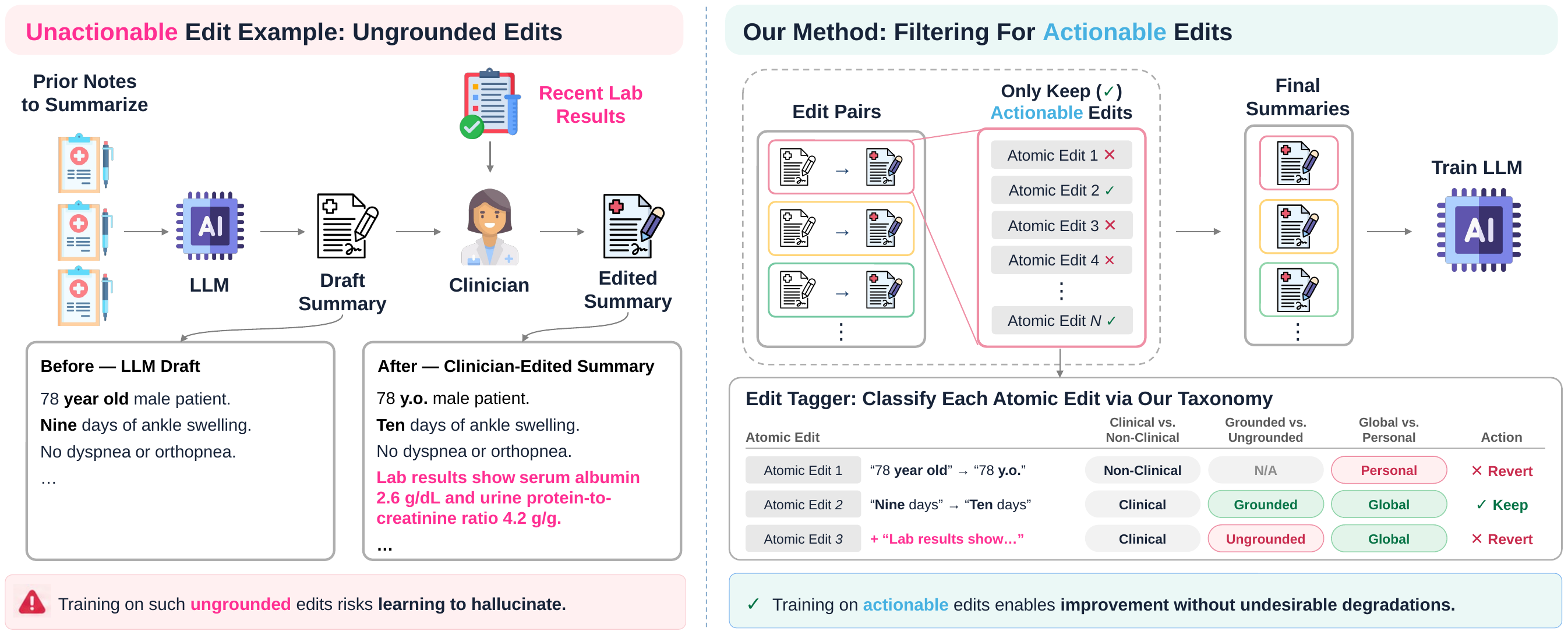}
    \vspace{-15pt}
    \caption{
    We propose a method for identifying and filtering for \textit{actionable} edits, i.e., edits that reveal gaps in model capabilities and carry rich supervision signal. 
    \textbf{(Left)} An example of an \textit{unactionable} edit is an \textit{ungrounded} edit, 
    where a clinician has access to information (e.g., recent lab results) that they edit into the final document, but which is \textit{not present} in the inputs to the LLM. 
    Example text is illustrative.
    Training an LLM to generate the edited summary from prior notes alone could risk increasing hallucinations (e.g., adding plausible but made-up lab results).
    \textbf{(Right)} An overview of our filtering method. From each (pre-edit, post-edit) summary pair, we extract all \textit{atomic edits} (Definition~\ref{def:atomic_edit}), categorize them based on our proposed taxonomy (Section~\ref{sec:taxonomy}), and revert edits that are not actionable (Section~\ref{sec:filtering}). 
    The resulting \textit{filtered} post-edit summaries are then used for model training, allowing improvements in draft quality without undesirable degradations (e.g., hallucinations).
    }
    \label{fig:conceptual}
    \vspace{-15pt}
\end{figure}%

However, there are good reasons to be cautious. 
First, a clinician's edit may be informed by information that was not available to the model, such as undocumented clinical observations or unrecorded conversations with the patient or other members of the care team (Figure~\ref{fig:conceptual}, left).
Training directly on such edits risks \textit{learning to hallucinate}, encouraging the model to emit outputs 
that are not grounded in its inputs. 
This problem is not hypothetical: 
for instance, more than 45\% of clinical notes, drafted by an ambient medical scribing product and edited by a clinician user, from a proprietary corpus of 30,000 de-identified and anonymized encounters are estimated to contain \textit{at least one ungrounded edit} (Section~\ref{sec:setting}, Appendix~\ref{appendix:ungrounded-prevalence}). 
Second, edits can be heterogeneous along certain dimensions (e.g., style or format) 
where the desired behavior of a system should be predictable. 
However, studying these challenges is hindered by a lack of public datasets that contain such post-deployment data (i.e., draft $\rightarrow$ edit data), despite the growing prevalence of such settings in real-world clinical workflows.

In this paper, we explore this problem setting
in the context of clinician edits to LLM-drafted discharge summaries.
To address the absence of publicly available clinician-edit datasets, 
we construct \dataset{}, 
a semi-synthetic dataset based on MIMIC-III~\citep{mimic-iii},
with LLM-drafted discharge summaries 
and their \textit{post-edit} variants (Figure~\ref{fig:dataset-construction}; Section~\ref{sec:dataset}).
The edits in our dataset are produced by grafting synthetic edits
onto LLM-drafted summaries, 
leveraging clinician-written discharge summaries for content. 
To ground the edits in real clinician behavior,
we design a taxonomy of edit types based on analysis of a 
proprietary dataset of 30,000 de-identified and anonymized encounters
from real-world deployment of an ambient medical scribing product (Section~\ref{sec:taxonomy_overall}).
The taxonomy of edit types and the style of these edits
inform the design of our LLM-based edit simulator. 
We also collect expert annotations to review the quality of the edits (Section~\ref{sec:edit-annotation}).

Using~\dataset{}, we demonstrate the challenges in directly training on raw edits and propose a taxonomy-based filtering approach that extracts edits that provide actionable supervision 
(Figure~\ref{fig:conceptual}, right). 
We find that naive fine-tuning on raw post-edit summaries improves draft quality over that of the base model on several metrics that measure alignment to clinician-written reference summaries, but at the cost of introducing ungrounded and clinically unsubstantiated content (Section~\ref{sec:result-naive}). 
In contrast, when fine-tuning on post-edit summaries after filtering out ungrounded and ``personal'' edits (i.e., reflecting local clinician preferences), 
we retain much of the performance gains observed from training on raw edit data, \textit{while largely preserving groundedness} (Section~\ref{sec:result-filtering}).

Overall, our work establishes a framework 
for studying clinician feedback through edits,
highlighting both its promise and pitfalls. 
Our main contributions are summarized as follows:
\begin{enumerate}[topsep=0.1ex,itemsep=0.1ex]
    \item We develop a taxonomy of clinician edits to LLM-generated clinical notes, capturing 
    how clinicians refine LLM outputs in practice (Section~\ref{sec:taxonomy_overall}). 
    \item We use our edit taxonomy to propose a method for identifying and categorizing edits, which enables scalable extraction of \textit{actionable feedback} from raw edit data (Section~\ref{sec:edit-tagger}).
    \item We introduce \dataset{}, a dataset of semi-synthetic edit pairs for the task of discharge summary generation (Section~\ref{sec:dataset}), 
    and use it as a testbed for studying how to effectively leverage noisy edit data for improving model capabilities (Section~\ref{sec:experiments}).
    \item We find that, on our dataset, (i) naively training on post-edit outputs can induce undesirable behavior, such as generating ungrounded content (Section~\ref{sec:result-naive}), and that (ii) \textit{filtering for actionable edits} is crucial to learn from edits without degradation in behavior
    (Section~\ref{sec:result-filtering}). 
\end{enumerate}
More broadly, our findings suggest that realizing the promise of post-deployment edit data as effective supervision signal warrants care in distinguishing those that are \textit{actionable} from others. 

\section{Related Works}
\label{sec:related-works}
\vspace{-7pt}
\subsection{Clinician Edits to Clinical Documentation}
\label{sec:edits-to-clinical-documents}
\vspace{-5pt}

One widely-adopted application of LLMs in healthcare is the use of ambient medical scribes that capture the audio of clinician-patient conversations and produce a first draft of clinical notes, which a clinician then reviews and edits before signing it into the EHR \citep{tierney2024-10,lukac2025-76,afshar2025-66}. 
The clinician edits naturally accumulate as a byproduct of routine documentation, motivating a nascent line of work that 
studies how clinicians make revisions, primarily through descriptive and qualitative analyses. 
Most such works focus on analyzing the \textit{frequency} of edits across specialties and visit types, 
while categorizing each edit based on the \textit{content} it modifies (e.g., medication, diagnosis, social history) with validation against human-annotated corpora \citep{cho2026consumer,guo2026clinicians,guo2026understanding}. 
Other works explore using lexical, semantic, and clinical edit distance metrics as coarse proxies for tracking the presence of clinically meaningful edits 
but do not characterize the nature of the specific content being modified \citep{clinician-edit-monitoring}. 
Beyond edit-based analyses, other works qualitatively explore clinician perspectives on ambient AI scribes through semi-structured interviews and thematic analyses \citep{van2026listening}. 
Collectively, these works establish that clinician edits encode systematic, clinically meaningful signal, 
but are limited to descriptive analyses on proprietary datasets, 
leaving the question of \textit{how one should leverage edits for targeted model interventions unanswered}.

\subsection{Learning from (Noisy) Edits}
\label{sec:learning-from-edits}
\vspace{-5pt}

\citet{Gao2024AligningLA} study learning from interactive user edits to LLM outputs in tasks such as news article summarization and email writing, adapting to latent user preferences inferrable from the edits. 
Unlike our setting, their setup assumes repeated interactions with a single user 
over multiple rounds, 
and their experiments rely on fully simulated edits that are not informed by real-world user-editing behavior. 
\citet{misra2026principledfinetuningllmsuseredits} adopt the setting of \citet{Gao2024AligningLA} and study the effectiveness and sample efficiency of different strategies for fine-tuning on edit data (e.g., supervised fine-tuning on post-edit outputs, preference learning~\citep{dpo} on (pre-edit, post-edit) output pairs). 
A distinct thread on learning from noisy user feedback considers noise in pairwise preference data, but models noise limited to label-flipping in pairwise comparisons \citep{im2026how}. 
In the clinical domain, \citet{yao2023improving} leverage token-level alignment between LLM drafts and clinician-edited summaries to improve patient-visit summary generation. 
However, they assume that all clinician edits
are suitable as ground-truth supervision, which our findings warrant caution against.

\section{Constructing a Taxonomy of Clinician Edits}
\label{sec:taxonomy_overall}
\vspace{-7pt}
\subsection{Analysis of Real-World Clinician Edit Data}
\label{sec:setting}
\vspace{-5pt}

Given the wide adoption of ambient scribing systems in healthcare (Section~\ref{sec:edits-to-clinical-documents}), we use the edits made to outputs of an ambient medical scribing product to study how clinicians edit LLM-generated clinical documentation and inform our taxonomy of clinician edits in Section~\ref{sec:taxonomy}. 
Our qualitative exploration focuses on edits to the Assessment and Plan (A\&P) section of a clinical SOAP note: a structured method of documentation used to record patient information in a chart. The A\&P contains the clinician's professional analysis or diagnosis and synthesizes the subjective and objective data into a working clinical impression which includes a plan with next steps for treatment. Our dataset contains 30,000 de-identified and anonymized encounters spanning more than 50 specialties from multiple institutions across the United States between 2023 and 2026.

As textual differences between the pre-edit and post-edit drafts of a de-identified and anonymized A\&P can span multiple distinct changes made by a clinician, we first extract from each (pre-edit, post-edit) pair a list of \textit{atomic edits}, which we define as follows: 

\begin{definition}[Atomic Edit]
\label{def:atomic_edit}
  Given a draft and its edited version, an \textbf{\textit{atomic edit}} is the smallest unit of textual change attributable to \textit{a single underlying clinician intent}. 
  All changes driven by the same reason are part of one atomic edit; changes driven by distinct reasons are separate atomic edits.
\end{definition}

\vspace{-5pt}

We implement the atomic-edit extraction step with GPT-4o~\citep{openai2024gpt4ocard}, which also outputs structured tags (e.g., factual error, omission) and descriptions of the clinician intent for each edit. 
Then, 
we obtain text descriptions of salient edit patterns via (i) topic modeling and (ii) sparse-autoencoder~\citep[SAE;][]{sae}-based feature discovery. 
For (i), we apply the BERTopic pipeline~\citep{grootendorst2022bertopic} to sentence embeddings of clinician-intent descriptions to identify clusters of atomic edits that reflect similar intentions. 
For (ii), we train an SAE on sentence embeddings of the clinician-intent descriptions and use the top-activating examples to characterize each learned feature.  
We observe that using the clinician-intent descriptions instead of the raw pre/post-edit notes is more effective in isolating the associated semantic change from other surface-level changes within each edit pair. 
We prompt GPT-5-mini~\citep{gpt-5} to obtain text descriptions of the atomic edit clusters from (i) and the learned SAE features from (ii). 
We incorporate our high-level findings from this process into constructing our taxonomy in Section~\ref{sec:taxonomy}. 

\subsection{Taxonomy of Clinician Edits to Clinical Documentation}
\label{sec:taxonomy} 
\vspace{-5pt}

We construct a taxonomy that characterizes clinician edits to LLM-generated clinical documentation along three axes, incorporating insights from (i) our exploration 
of clinician edit data from an ambient medical scribing vendor (Section \ref{sec:setting}) and (ii) prior studies of clinician-editing behavior in the literature \citep{guo2026clinicians, cho2026consumer}. 
We design our taxonomy to 
capture the heterogeneity of edits observed throughout deployment 
and provide a principled basis for identifying edits suitable as training signal. 
The three axes in our taxonomy each capture: what content is being edited (\textit{Clinical vs.~Non-Clinical}), whether the edit is supported by the input context (\textit{Grounded vs.~Ungrounded}), and whether the edit reflects local preferences 
(\textit{Global vs.~Personal}).

\vspace{-7pt}
\paragraph{Clinical vs.~Non-Clinical Edits.} We distinguish between \textit{clinical} and \textit{non-clinical} edits based on whether they modify clinical content in the pre-edit draft. 
This distinction matters because the two types of edits carry fundamentally different implications for model improvement: 
clinical edits often encode generalizable signal about what the model should produce, whereas non-clinical edits may mainly reflect personal preferences. 
We further categorize each clinical edit as one of five subtypes, based on the underlying clinician intent: 
factual error correction, omission correction, condensation, specificity adjustment, and confidence-level adjustment (definitions in Appendix~\ref{appendix:clinical_edits}). 
Non-clinical edits, by contrast, leave clinical content intact and instead modify formatting, organization, or stylistic properties (e.g., bullet points vs. free-form prose, abbreviation style of clinical terminology). 

\vspace{-7pt}
\paragraph{Grounded vs.~Ungrounded Edits.} For \textit{clinical} edits, we further distinguish between edits whose resulting changes in clinical content are supported by the context provided to the model (e.g., prior notes) and others. 
A clinical edit is \textit{grounded} if the associated change can be explained by information either present or absent from the supporting evidence. 
For example, an omission correction is grounded when the added information appears in the input context (e.g., prior notes) but was missed by the model, 
and a factual error correction is grounded when the edited version aligns with---and the original conflicted with---clinical details present in the input context.  
A clinical edit is \textit{ungrounded} when no such evidence-based explanation exists: the edit reflects information, judgment, or preference exogenous to the model's inputs. 
These edits are not necessarily inaccurate, as they may reflect privileged knowledge or context the model cannot access, and are prevalent in real clinician-edit data (Appendix~\ref{appendix:ungrounded-prevalence}). 
However, treating them as supervision targets risks training the model to confabulate beyond its evidence, since the ``correct'' output is not derivable from the model's inputs. 

\vspace{-7pt}
\paragraph{Global vs.~Personal Edits.} 
This third axis introduces another level of actionability. 
Among edits suitable as learnable signal, some reflect the population as a whole while others reflect feedback from particular subgroups or individuals. Personal edits reflect stylistic preferences that vary across clinicians: e.g., bullet points or free-form prose, expanded or contracted abbreviations for medical terminology, and how to order and group problems within the assessment. Such edits are most often non-clinical, but the categories are not coextensive: confidence-level adjustments and condensation patterns, for example, could reflect personal practice style despite affecting clinical content. 

\section{\textsc{MIMIC-III-Ext-BHC-Edit}: A Semi-Synthetic Clinician Edit Dataset for Brief Hospital Course Generation}
\label{sec:dataset}
\vspace{-7pt}
\begin{figure*}[t!]
    \centering
    \includegraphics[width=0.95\linewidth]{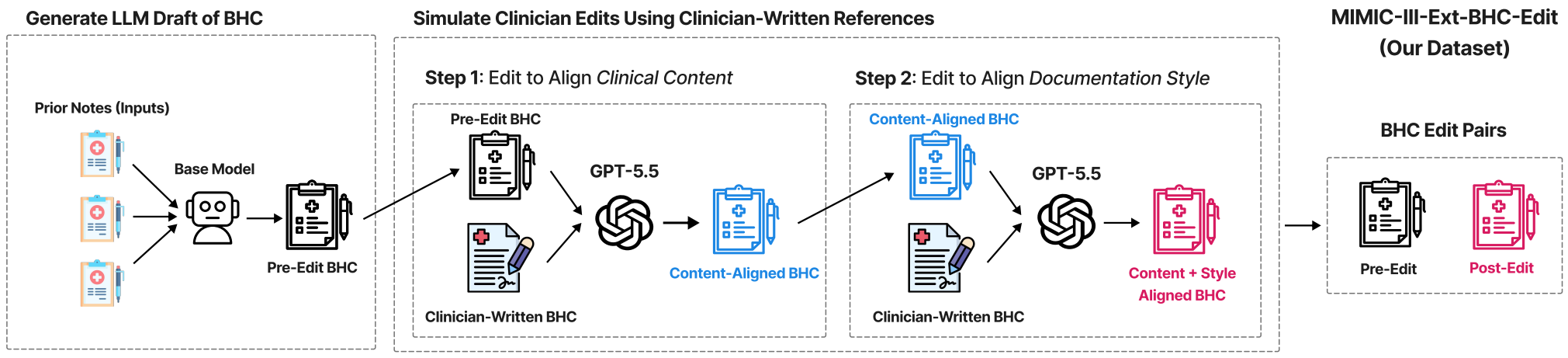}
    \vspace{-6pt}
    \caption{Overview of our generation process for \dataset{}, a semi-synthetic dataset of edit pairs for the task of generating brief hospital course (BHC) summaries (Section~\ref{sec:dataset}).}
    \label{fig:dataset-construction}
    \vspace{-5pt}
\end{figure*}

We introduce \dataset{}, a semi-synthetic dataset based on MIMIC-III consisting of LLM-drafted discharge summaries (i.e., the \textit{pre-edit} notes) and their \textit{post-edit} versions, 
simulated 
based on the clinician-written summary available for each admission (Figure~\ref{fig:dataset-construction}). 
Following prior works on discharge summary generation \citep{whats-in-a-summary,discharge-me,verifact}, we focus on the BHC section, which details the key clinical events, diagnoses, and treatments during a patient's stay in the hospital.
As inputs for the task, we provide all prior notes available for the patient up to discharge time (e.g., nursing charts, chest X-ray reports). 
We extract a total of 5,821 admissions from MIMIC-III, using the cohort selection criteria by 
\citet{verifact} (Appendix~\ref{appendix:mimic-cohort-selection}). 

\vspace{-7pt}
\paragraph{Generating the pre-edit BHCs.} 
To generate the initial BHC draft for each admission, we zero-shot prompt three LLMs with the pre-discharge EHR notes as context (Figure~\ref{fig:dataset-construction}, left): Gemma-3-4B-IT \citep{gemma3}, Qwen-2.5-7B-Instruct \citep{qwen25}, and Llama-3.1-8B-Instruct \citep{llama-3} (see Appendix~\ref{appendix:pre-edit-prompt} for the prompting details). 
We select these models as they support long-context inputs (up to 128K), cover various model architectures and sizes, and are open-weight models that can be fine-tuned.  
We simulate the clinician edits on top of the zero-shot outputs, which provide the supervision signal for fine-tuning each model for BHC generation. 

\vspace{-7pt}
\paragraph{Generating the post-edit BHCs.} From the pre-edit BHCs, we generate their \textit{post-edit} versions, using the clinician-written reference BHC for each admission as a proxy for how a clinician would have edited the LLM draft (Figure~\ref{fig:dataset-construction}, middle). 
We perform this in two steps: \textbf{Step 1} (\textit{clinical content} alignment) updates the clinical content of the pre-edit BHC to match that of the reference; and \textbf{Step 2} (\textit{documentation style} alignment) updates the style (e.g., problem-based organization, use of shorthands) of the content-aligned BHC to match that of the reference. 
We implement each step with GPT-5.5 \citep{gpt-55} in compliance 
with the PhysioNet Credentialed Health Data License.

In \textbf{Step 1}, we provide an LLM with the pre-edit and reference BHCs for each admission and prompt it to add or remove clinical details such that the two BHCs match in clinical content, while preserving the documentation style (e.g., paragraph structure, prose style). 
A \textit{clinical detail} is a discrete piece of clinical information, such as a single medication with its dosage or a single diagnosis with its supporting rationale. 
As these edits are all \textit{clinical} based on our taxonomy (Section~\ref{sec:taxonomy}), we simulate each edit as 
one of six possible operations each aligned with a clinical-edit subtype: 
factual error correction, omission correction, condensation, specificity adjustment, confidence level adjustment, and retain. 
We provide full details on per-operation rules in Appendix \ref{appendix:step1-details}. 
As the pre-edit drafts are generally much shorter than the clinician-written references, frequent omission corrections (around 23--32\% of edit operations; Tables~\ref{tab:step_1_edit_ops_gemma_3_4b_it}--\ref{tab:step_1_edit_ops_llama_3_1_8b_instruct}) tend to make the post-edit BHCs longer (Table~\ref{tab:edit_pair_stats}). 

In \textbf{Step 2}, 
we provide an LLM with the content-aligned BHC from \textbf{Step 1} and the reference BHC and prompt it to revise the former to match the structural and stylistic conventions of the latter, 
\textit{without altering its clinical content}, 
in five aspects: 
organization by problem vs.~chronology; bullet points vs.~paragraphs; shorthands (e.g., ``year old'' vs.~``y.o.''); 
fragmented vs.~complete sentences; and the ordering and grouping of problems.
Since clinicians often vary in documentation preferences, \textbf{Step 2} can be viewed as \textit{personalizing} the content-aligned BHC to the clinician who wrote the reference. 

\subsection{Expert Validation of the Simulated Edits}
\label{sec:edit-annotation}
\vspace{-5pt}

To ensure that the simulated edits result in BHCs that (1) read as coherent and professionally written documentation; and (2) are well-aligned with clinician-written references in clinical content, we conduct an expert annotation study with a pool of 11 nurse practitioners (NPs) with hospitalist backgrounds to assess each aspect. 
Starting from the pool of (pre-edit, \textbf{Step 2}) BHC pairs generated for all three base LLMs, we randomly sample 25 edit pairs for assessing both (1) and (2). 
From each edit pair, we extract a list of all \textit{atomic edits} (Definition~\ref{def:atomic_edit}) and randomly select one for annotation. 
Alongside each edit being annotated, we provide the NPs with the corresponding clinician-written reference BHC, and the pre- and post-edit BHCs to provide relevant context. 
For each example, we obtain annotations from three NPs (75 total judgments over 22.5 hours of annotation effort) and acquire a consensus label via majority voting. 
We include the full annotation guide in Appendix~\ref{appendix:edit-annotation}. 

\vspace{-7pt}
\paragraph{Task 1: Coherence \& Documentation Practice.} To assess whether each edit produces coherent and professionally written changes that practitioners would reasonably make, 
we ask each annotator to provide one of three ratings: \textit{Reasonable}, \textit{Unreasonable}, and \textit{Unsure} (detailed definitions of the ratings in Appendix~\ref{appendix:edit-annotation}). All 25 examples were labeled as \textit{Reasonable}, with unanimous agreement on 88\% (22/25) and majority agreement on 12\% (3/25) of cases (Table~\ref{tab:task1-expert-validation}, top row). 

\vspace{-7pt}
\paragraph{Task 2: Clinical Content Alignment to the Reference.} To assess whether each edit only involves changes that agree with the reference in clinical content, we ask each annotator to provide one of four ratings: \textit{Aligned}, \textit{Not Aligned}, \textit{Unsure}, or \textit{N/A}. 84\% (21/25) of examples were labeled as \textit{Aligned}, while 16\% (4/25) were labeled as \textit{N/A} (i.e., edit does not change clinical content), with unanimous agreement on 80\% (20/25) and majority agreement on 20\% (5/25) of cases (Table~\ref{tab:task1-expert-validation}, bottom row). 

Overall, we observe that experts generally find our simulated edits to reflect changes that practitioners would reasonably make and to be well-aligned with clinician-written references.
We do not claim that our simulated edits in \dataset{} exactly reproduce how clinicians edit LLM drafts under real-world workflow constraints. 
Rather, our dataset provides a controlled testbed in which grounded, ungrounded, global, and personal edits can be isolated. 
Every added or corrected detail in the post-edit BHCs comes from a real clinician-written BHC; thus, even ungrounded content reflects information clinicians actually document rather than arbitrary insertions.

\section{Identifying and Isolating Actionable Edits for Training}
\label{sec:edit-tagger}
\vspace{-7pt}
\begin{table*}[!t]
  \centering
  \small
  \caption{Expert validation of the simulated edits in \dataset{} (Section~\ref{sec:edit-annotation}).
  }
  \vspace{-5pt}
  \resizebox{0.8\textwidth}{!}{
  \begin{tabular}{llcccc}
  \toprule
  \multirow{2}{*}{Annotation Task} & \multirow{2}{*}{Expert Rating} & \multicolumn{2}{c}{Examples} & \multicolumn{2}{c}{Inter-Annotator Agreement} \\
  \cmidrule(lr){3-4} \cmidrule(lr){5-6}
   & & \% & $n$ & Unanimous & Majority \\
  \midrule
  \multirow{4}{*}{\shortstack[l]{\textbf{Task 1: Coherence \&}\\\textbf{Documentation Practice}}} & \textbf{Reasonable} & \textbf{100\%} & 25 & 22 & 3 \\
   & Unreasonable & 0\% & 0 & 0 & 0 \\
   & Unsure & 0\% & 0 & 0 & 0 \\
  \cmidrule(lr){2-6}
   & \textit{All} & 100\% & 25 & \textbf{88\%} (22/25) & \textbf{12\%} (3/25) \\
  \midrule
  \multirow{5}{*}{\shortstack[l]{\textbf{Task 2: Clinical Content}\\\textbf{Alignment to Reference}}} & \textbf{Aligned} & \textbf{84\%} & 21 & 19 & 2 \\
   & Not Aligned & 0\% & 0 & 0 & 0 \\
   & Unsure & 0\% & 0 & 0 & 0 \\
   & N/A & 16\% & 4 & 1 & 3 \\
  \cmidrule(lr){2-6}
   & \textit{All} & 100\% & 25 & \textbf{80\%} (20/25) & \textbf{20\%} (5/25) \\
  \bottomrule
  \end{tabular}
  }
  \label{tab:task1-expert-validation}
  \vspace{-5pt}
\end{table*}

To identify the \textit{actionable} edits to use as supervision signal, 
we develop a GPT-5.5-based \textit{edit tagger} that takes an edit pair as input and produces structured per-edit annotations aligned with our taxonomy. 
Given (i) the pre-edit BHC, 
(ii) the post-edit BHC, and (iii) all pre-discharge EHR notes for a given admission, 
the edit tagger extracts all atomic edits (Definition~\ref{def:atomic_edit}) and classifies each along the three axes of our taxonomy (full details in Appendix~\ref{appendix:tagging_prompt}). 
Table~\ref{tab:edit-distribution} shows the distribution of edit types with the \textbf{Step 1} and \textbf{Step 2} outputs as the post-edit BHC, as determined by the edit tagger. 

We also assess the accuracy of the edit tagger judgments along all three axes via expert annotations (full details in Appendix~\ref{appendix:tagger-eval}). 
For the \textit{Clinical vs.~Non-Clinical} and \textit{Global vs.~Personal} axes, we compare the edit tagger outputs to expert annotations that we collect from a pool of 10 NPs with hospitalist backgrounds, on 25 randomly sampled atomic edits. 
For each edit, we obtain annotations from three NPs (75 total judgments over 22.5 hours of annotation effort). 
For the \textit{Grounded vs.~Ungrounded} axis, we compare the edit tagger outputs for 150 randomly sampled atomic edits (for each base model in \dataset{}; Section~\ref{sec:dataset}) to the clinician annotations from \verifactbhc{} \citep{verifact}, which assess the groundedness of each ``claim'' from a reference BHC. 
As our atomic edits and the \verifactbhc{} claims do not completely align, we only review edits whose text spans align with at least one claim to obtain the clinician ground-truth. 

Our edit tagger agrees the most with experts on the \textit{Clinical vs.~Non-Clinical} (75\% = 18/24, excluding one case labeled \textit{Unsure}) and \textit{Grounded vs.~Ungrounded} (e.g., 78\% = 117/150 for Gemma-3-4B-IT)\footnote{See Appendix \ref{appendix:tagger-eval} for the agreement rates for Qwen-2.5-7B-Instruct and Llama-3.1-8B-Instruct BHCs.} axes. 
Experts agree unanimously with each other on 84\% (21/25) and 63.1\% (149/236)\footnote{Inter-annotator agreement for Gemma-3-4B-IT. The 236 figure is the number of \verifactbhc{} claims that overlap with the 150 atomic edits reviewed. See Appendix~\ref{appendix:tagger-eval} for results on the other base models.} of cases, respectively.
Agreement between the edit tagger and experts is lowest on \textit{Global vs.~Personal} (52\% = 13/25), which is also more subjective for experts than \textit{Clinical vs.~Non-Clinical}: experts agree unanimously on only 64\% (16/25) of the same edits. 
We thus treat \textit{Global vs.~Personal} tags as noisier signal than those on the other two axes. 

\vspace{-7pt}
\paragraph{Learning from \textit{actionable} edits.}
\label{sec:filtering}
After extracting and tagging the atomic edits from each (pre-edit, post-edit) BHC pair, we construct various \textit{filtered} training sets, where we selectively \textit{revert} edits of specific types. 
Given each pair and filter, we update the post-edit BHC by undoing every
atomic edit whose tags are excluded by the filter: 
reverting insertions, restoring deletions, and substituting the original draft span back in for substitutions 
(Appendix~\ref{appendix:filtering}). 
This yields a family of training targets that share the same input but differ in which edits they treat as supervision signal. 
We define two filters that progressively restrict the supervision to increasingly conservative subsets of edits:

\begin{itemize}[topsep=0.1ex,itemsep=0.1ex]
\item \textbf{\textit{Grounded} edits}: Removes \textit{clinical} edits that are \textit{ungrounded} in the inputs, restricting supervision to other edits explicitly substantiated by the input evidence available to the model.
\item \textbf{\textit{Grounded + Global} edits}: Additionally removes \textit{personal} edits, restricting to edits that improve general model capabilities
rather than reflecting individual clinician preferences.
\end{itemize}
We focus on removing ungrounded and personal edits, as they are least likely to provide generalizable supervision: the former are unsupported by the input context; the latter reflect local preferences.

\section{Experiments}
\label{sec:experiments}
\vspace{-7pt}
We validate our method for extracting and learning from actionable clinician edits (Section~\ref{sec:filtering}) using \textsc{MIMIC-III-Ext-BHC-Edit} (Section~\ref{sec:dataset}). 
In particular, we focus on improving the clinical utility of the generated BHC summary \textit{without losing groundedness to the prior notes provided as input}. 
We also assume a setting where clinicians edit LLM drafts to complete documentation tasks, and we thus do not have access to the ``oracle'' summaries that clinicians would have written on their own. 

Starting from Gemma-3-4B-IT, Qwen-2.5-7B-Instruct, and Llama-3.1-8B-Instruct---the three models whose zero-shot BHC generations serve as the \textit{pre-edit} BHCs in our \dataset{} dataset (Section~\ref{sec:dataset})---we perform supervised fine-tuning (SFT) with low-rank adapters \citep[LoRA;][]{lora}, using each model's simulated \textit{post-edit} BHCs as the ground-truth. 
We compare between SFT on the 
\textit{Grounded} and \textit{Grounded + Global}
edit pairs resulting from our filtering method (Section~\ref{sec:filtering})
vs.~SFT on the raw, unfiltered post-edit BHCs. 
We also compare with using the original clinician-written BHCs for SFT (where one assumes access to fully human-authored notes), as well as on intermediate outputs from post-edit simulation (\textbf{Step 1}; Section~\ref{sec:dataset}) (see Appendix~\ref{appendix:sft-clinician-step1}).

We evaluate on two test sets---one based on the 100 admissions that overlap with \verifactbhc{} \citep{verifact}, and another obtained from taking a 90--5--5 train, validation, and test split by patient on the remaining 5,721 admissions from our dataset. 
We curate the former as a \textit{grounded} test set, using the human annotations from \verifactbhc{} of whether each atomic claim from the reference (clinician-written) BHC is substantiated by the prior notes (Appendix~\ref{appendix:grounded-test-set}). 
The \textit{grounded} test set is considered to avoid rewarding models for generating ungrounded clinical information. 

\vspace{-7pt}
\paragraph{Evaluation metrics.} We assess the quality of model-generated BHCs with the following metrics:
(i) \textit{clinical claim/term-matching} metrics: DocLens-Precision/Recall~\citep{doclens} and MEDCON \citep{aci-bench}; 
(ii) \textit{clinical factuality} metrics: MedScore \citep{medscore}, DocLens-Attribution~\citep{doclens}, and PDSQI-9 \citep{pdsqi-9}; 
and 
(iii) natural language generation (NLG) metrics: BERTScore \citep{zhang2020bertscoreevaluatingtextgeneration}, ROUGE-L \citep{rouge}, and BLEU-2 \citep{bleu}.
The metrics in (i) assess whether the clinical claims/terms in each model-generated BHC match those in the reference BHC, 
measuring alignment to what a clinician would document. 
The metrics in (ii) assess whether the claims in each model-generated BHC are supported by the input prior notes, measuring faithfulness to the input. 
DocLens-Attribution is an adaptation of the ``Citation Recall'' metric in \citet{doclens}, checking whether each claim from the model-generated BHC is entailed by the prior notes instead of a model-generated citation. 
For PDSQI-9, we focus on its ``accurate'' score (1--5), which measures factuality at the summary level. 
As other metrics range between 0 and 1, we shift and scale PDSQI-9 to the same range. 
We compute the reference-based metrics in (i) and (iii) on the \textit{grounded} test set (\verifactbhc{}), and all other metrics on examples from both the \verifactbhc{} and \dataset{} test sets.

\section{Results}
\label{sec:results}
\vspace{-7pt}
We present the main findings from our experiments (Section~\ref{sec:experiments}), focusing on the \textit{clinical claim/term-matching} and \textit{clinical factuality} metrics. 
Overall, we find that naive SFT on raw edit data improves BHC generation but degrades groundedness in the input prior notes (Section~\ref{sec:result-naive}), 
while filtering out ungrounded edits beforehand enables performance gains without compromising groundedness (Section~\ref{sec:result-filtering}). 
We include the results from NLG-based evaluation and additional analyses in Appendix~\ref{appendix:results}, where we show that similar trends hold for SFT on the original clinician-written references available from MIMIC-III (``oracle'' supervision) or the \textit{intermediate} post-edit BHCs (\textbf{Step 1}; Section~\ref{sec:dataset}).

\begin{figure*}[t!]
\centering
\begin{tabular}{c@{}c}
    \begin{subfigure}{0.03\linewidth}
        \makebox[\linewidth]{\raisebox{55pt}{{(a)}}} % 65
    \end{subfigure} &
    \begin{subfigure}{0.85\linewidth} % 0.85
        \includegraphics[width=\linewidth]{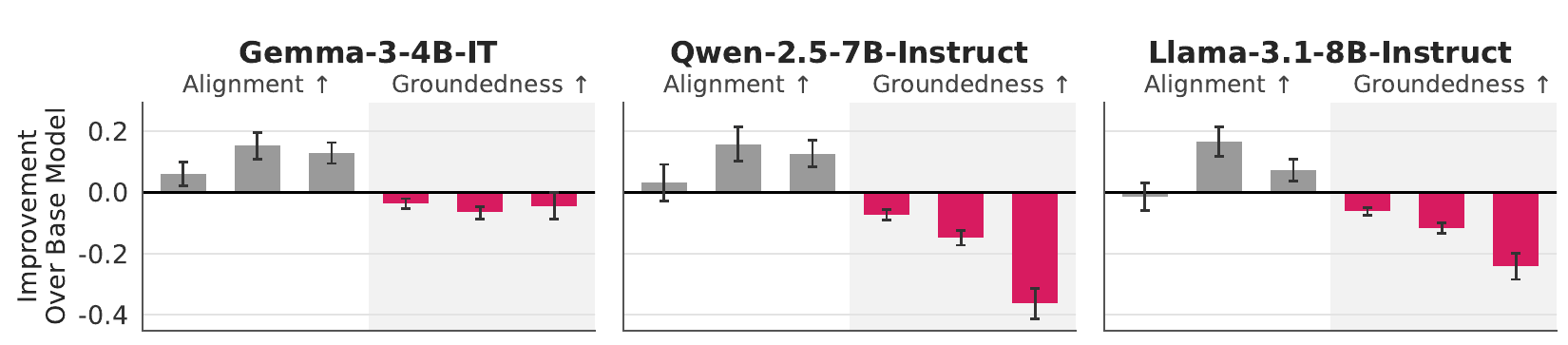}
    \end{subfigure} \\
    \begin{subfigure}{0.03\linewidth}
        \makebox[\linewidth]{\raisebox{55pt}{{(b)}}} % 55
    \end{subfigure} &
    \begin{subfigure}{0.85\linewidth} % 0.85
        \includegraphics[width=\linewidth]{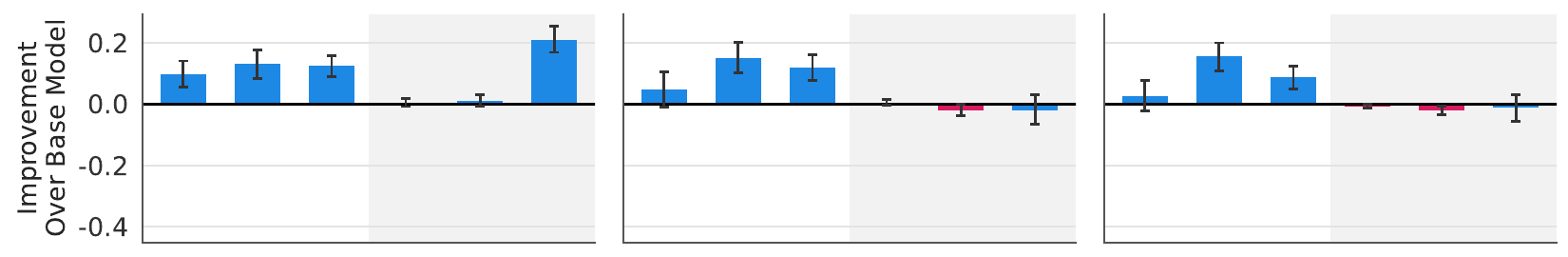}
    \end{subfigure} \\
    \begin{subfigure}{0.03\linewidth}
        \makebox[\linewidth]{\raisebox{90pt}{{(c)}}} % 55
    \end{subfigure} &
    \begin{subfigure}{0.85\linewidth} % 0.85
        \includegraphics[width=\linewidth]{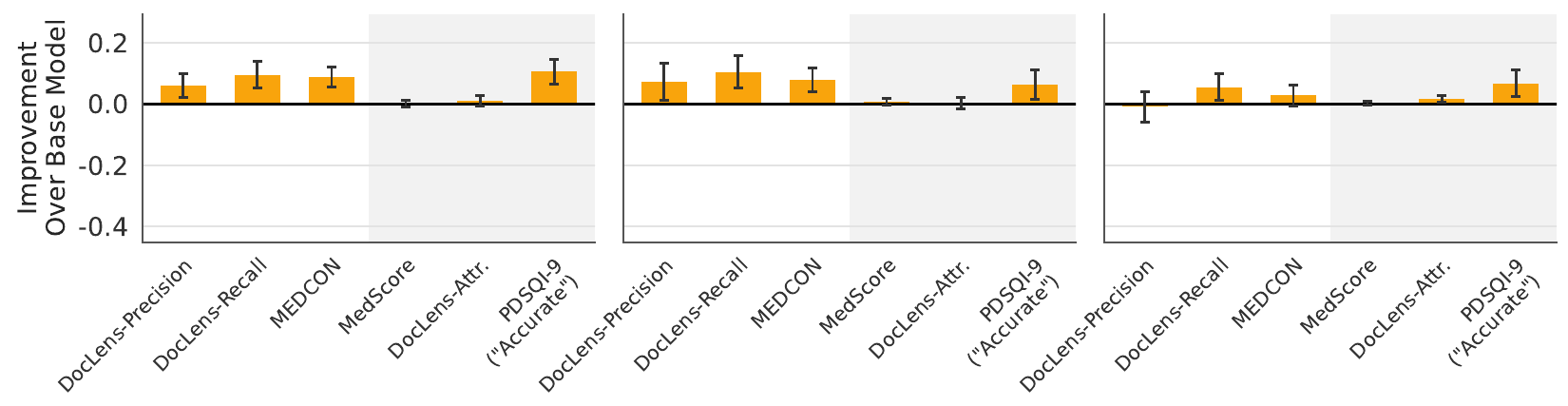}
    \end{subfigure}
\end{tabular}
\vspace{-10pt}
\caption{Naive fine-tuning on raw post-edit BHCs degrades groundedness in the input prior notes (Section~\ref{sec:result-naive}), while filtering for actionable edits beforehand improves alignment to clinician-written references and \textit{largely preserves groundedness} (Section~\ref{sec:result-filtering}). 
We report improvements over the base model in the 
clinical claim/term-matching metrics and 
the groundedness metrics, 
after SFT on the (a) \textcolor{raw}{raw}, 
(b) \textcolor{grounded}{\textit{Grounded}}, and (c) \textcolor{groundedglobal}{\textit{Grounded + Global}} post-edit BHCs. 
In each panel, the left three metrics measure alignment and the shaded right three metrics measure groundedness. 
Note that the metrics are not necessarily comparable on the same scale, but the point is in their  directionality. 
Error bars indicate the 95\% bootstrap CIs, computed by resampling the test set with replacement 10,000 times, and \textcolor{degraded}{red} bars indicate statistically significant degradations relative to the base model. 
}
\label{fig:main-figure}
\vspace{-10pt}
\end{figure*}

\subsection{Naive SFT on Raw Edits Improves Performance But Degrades Factuality}
\label{sec:result-naive}
\vspace{-5pt}

Across all three models, directly fine-tuning on the raw edit data generally improves alignment to clinician-written reference BHCs, suggesting that edits can provide effective supervision for model training (Figure~\ref{fig:main-figure}(a), left half of each panel). 
For instance, the average MEDCON score on the \textit{grounded} test set (Section~\ref{sec:experiments}) increases by +0.130 for Gemma-3-4B-IT (0.229 $\rightarrow$ 0.359), +0.127 for Qwen-2.5-7B-Instruct (0.253 $\rightarrow$ 0.380), and +0.073 for Llama-3.1-8B-Instruct (0.290 $\rightarrow$ 0.363), all of which are statistically significant improvements (i.e., 95\% bootstrap confidence intervals (CIs) in average improvement lie above 0). 
DocLens-Recall scores also increase significantly, indicating that the BHCs generated by the fine-tuned models provide better coverage of clinical details in the clinician-written reference, while maintaining similar DocLens-Precision scores as the base model.

However, these improvements come at the cost of groundedness in the input prior notes (Figure~\ref{fig:main-figure}(a), shaded right half of each panel). 
For example, the average PDSQI-9 ``accurate'' score (on the original scale of 1--5) drops by 0.18 for Gemma-3-4B-IT (2.32 $\rightarrow$ 2.14), 1.45 for Qwen-2.5-7B-Instruct (3.39 $\rightarrow$ 1.94), and 0.97 for Llama-3.1-8B-Instruct (2.92 $\rightarrow$ 1.95). 
In fact, all models show statistically significant degradations on MedScore, DocLens-Attribution, and PDSQI-9 (shown in the \textcolor{degraded}{red} bars). 
Notably, such degradation in groundedness is a direct consequence of treating the raw post-edit BHCs with high levels of ungroundedness as the ground-truth (Figure~\ref{fig:train-groundedness}, middle). 
Our results suggest that while training on raw edit data can help improve the general summarization capabilities of the model, it can also induce the model to generate content that is not faithful to the input context.

These results also suggest that relying on reference-based metrics for evaluating model-generated clinical summaries warrants caution. 
Even after removing any ungrounded details from the reference (e.g., \textit{grounded} test set), 
such metrics only indicate whether a summary sufficiently captures the key details that a clinician would highlight, 
and may fail to penalize cases where the generated details are consistent with the reference but not supported by the input prior notes. 
Therefore, it is crucial 
to assess reference-based agreement \textit{subject to the constraint that groundedness does not degrade}.

\subsection{SFT on Filtered Edits Improves Performance While Preserving Factuality}
\label{sec:result-filtering}
\vspace{-7pt}

After applying our \textit{Grounded} and \textit{Grounded + Global} filters to the BHC edit pairs (Section~\ref{sec:filtering}), SFT on the post-edit BHCs improves alignment to clinician-written references \textit{while generally preserving groundedness} (Figure~\ref{fig:main-figure}(b,c)). 
For instance, fine-tuning Gemma-3-4B-IT with the filtered data results in significant improvements on \textit{all} reference-based metrics (i.e., 95\% bootstrap CIs in average improvement lie above 0), with similar or better groundedness scores (notably, PDSQI-9) than those of the base model. 
Qwen-2.5-7B-Instruct and Llama-3.1-8B-Instruct also generally show improvements on DocLens-Recall and MEDCON, especially in the \textit{Grounded} filter setting, with little to no degradation in groundedness.
These trends arise from treating BHCs with high groundedness scores as the ground-truth, as enabled by our edit filters (Figure~\ref{fig:train-groundedness}, right). 
Varying the fraction of ungrounded edits removed shows that groundedness after SFT improves steadily as more such edits are filtered out, while reference-alignment gains remain largely unchanged (Appendix~\ref{appendix:dose-response}). 

Notably, the models trained with our filtered edit data retain groundedness \textit{while producing longer and more detailed BHCs} than the original base model, a setting where factuality is more challenging to achieve (Table~\ref{tab:bhc-len-all}). 
For instance, for Qwen-2.5-7B-Instruct, the average length of BHCs from the base model is only 109.3 words, while the average lengths of BHCs after SFT on the \textit{Grounded} and \textit{Grounded + Global} post-edit BHCs are 219.9 and 177.3 words, respectively. 
We observe similar trends for Gemma-3-4B-IT and Llama-3.1-8B-Instruct (Table~\ref{tab:bhc-len-all}, top and bottom rows). 
These results demonstrate the effectiveness of our edit tagger and filtering strategies (Section~\ref{sec:filtering}) in translating noisy edit data into effective supervision signal for model training. 

Meanwhile, although improvements are evident along certain axes, no single training setup dominates across all metrics. 
For example, the groundedness scores for Qwen-2.5-7B-Instruct and Llama-3.1-8B-Instruct are maximized by our most aggressive filter (\textit{Grounded} + \textit{Global}; Figure~\ref{fig:main-figure}(c)),  
while the reference-based metrics (MEDCON, DocLens) increase more when only filtering out ungrounded details (\textit{Grounded}; Figure~\ref{fig:main-figure}(b)), suggesting that some edits tagged as ``personal'' may still provide useful supervision (Section~\ref{sec:edit-tagger}). 
These results suggest that more sophisticated filtering  
strategies may be necessary to yield further gains and generalize across different edit types. 

\subsection{Expert Validation of Model-Generated BHC Groundedness}
\label{sec:expert-validation-groundedness}
\vspace{-7pt}
For further validation of our findings in Sections~\ref{sec:result-naive}--\ref{sec:result-filtering}, we obtain expert annotations from a pool of 8 NPs with hospitalist backgrounds on 10 admissions randomly selected from the test sets. 
For each admission, we ask each annotator to compare the three BHCs generated by the fine-tuned models (raw, \textit{Grounded}, \textit{Grounded + Global}) against the zero-shot-generated BHC from the base model in terms of their groundedness. 
For each pairwise comparison, annotators are asked to provide one of three ratings: 
\textit{A > B}, \textit{A < B}, \textit{A $\approx$ B},
where A and B each denote either the BHC generated by one of the three fine-tuned models or the zero-shot-generated BHC, shown in random order. 
We obtain 3 ratings per comparison, totaling $10 \times 3 \times 3 = 90$ judgments over 30 total hours of annotation effort. 

Comparing SFT (raw) to the base model, we find that the base-model BHCs are considered strictly better grounded in 80\% of admissions. 
In contrast, the BHCs generated by SFT (\textit{Grounded}) and SFT (\textit{Grounded + Global}) are both similar to or better than the base-model BHCs in 90\% (Better: 30\%, Equal: 60\%) and 60\% (Better: 50\%, Equal: 10\%) of admissions, respectively (Table~\ref{tab:task2-groundedness}). 
These results corroborate our finding that our filtering strategies help mitigate degradation in groundedness. 
We provide the remaining details (e.g., inter-annotator agreement) from this study in Appendix~\ref{appendix:groundedness-annotation}.

\section{Discussion and Conclusion}
\label{sec:discussion}
\vspace{-7pt}
In this paper, we studied how to leverage clinician-edit data for model training. 
Informed by real-world clinician-edit data, 
we constructed a taxonomy of clinician edits, 
distinguishing edits that are actionable (e.g., grounded, global) 
from others that are not (e.g., ungrounded, personal). 
We then introduced \dataset{}, a semi-synthetic edit dataset based on MIMIC-III, given the lack of publicly available edit datasets. 
On this dataset, we found that naively training on raw edits can degrade generation quality, and that targeted removal of edits with undesirable properties (e.g., ungrounded) can help 
translate edit data into effective supervision signal without clear degradations. 
The broader message of our work is that the promise of post-deployment feedback data for improving model capabilities depends on recognizing that not all feedback is actionable; methods to understand this distinction are a prerequisite for any reliable use of edit data as supervision and for models to continuously improve from real-world use at scale.

\vspace{-7pt}
\paragraph{Limitations.} We present our findings with the following caveats. 
First, our edits are semi-synthetic: while our expert annotations provide support for their quality (Section~\ref{sec:edit-annotation}), 
they may not capture every aspect of how clinicians edit LLM drafts in real clinical workflows. 
However, they provide a controlled testbed in which grounded, ungrounded, global, and personal edits can be isolated, with every added or corrected detail drawn from a real clinician-written BHC. 
Second, our edit simulation and filtering pipelines are based on a single LLM (GPT-5.5; Sections~\ref{sec:dataset}--\ref{sec:edit-tagger}). 
In ablations, we find that our main findings largely hold with Claude Opus 4.8~\citep{opus-4-8} used for filtering edits simulated by GPT-5.5 (Appendix~\ref{appendix:ablation-tagging-filtering}). 
We leave additional validation under alternative simulation and filtering setups to future work. 
Third, we assess the clinical quality of model-generated BHCs primarily with reference-based clinical claim/term-matching metrics (Section~\ref{sec:experiments}), which are imperfect proxies for clinical utility. 
Fourth, we largely treat personal edits as noisy supervision, and our tagger separates global from personal edits less reliably than along other axes (Section~\ref{sec:edit-tagger}).  Leveraging them for adaptation to local settings is another promising direction, which we leave to future work. 
Fifth, as MIMIC-III may not capture all information originally available to clinicians, some reference content may appear ungrounded only because supporting notes are missing, resulting in inflated ungroundedness levels in some cases. 
Lastly, we study a single clinical documentation type; analogous benchmarks for expert edits in other domains (e.g., law, science) are a natural extension.

\subsubsection*{Acknowledgments}
ZCL gratefully acknowledges NSF (IIS-2211955), Abridge AI, UPMC, Highmark Health, Ford Research, Mozilla, the PwC Center, Amazon AI, JP Morgan Chase, the Block Center, the Center for Machine Learning and Health, and the CMU Software Engineering Institute (SEI) via Department of Defense contract FA8702-15-D-0002, for the generous support of ACMI Lab's research. 
We also thank Alexandra Chouldechova for helpful feedback and comments on earlier versions of this work. 

\clearpage

\bibliography{refs}
\bibliographystyle{main}

\clearpage
\appendix

\renewcommand\thetable{A\arabic{table}}
\renewcommand\thefigure{A\arabic{figure}}
\setcounter{table}{0}
\setcounter{figure}{0}
\section{Additional Details on the Edit Taxonomy}
\label{appendix:taxonomy}
\subsection{Clinical Edit Subtypes}
\label{appendix:clinical_edits}

We further categorize \textit{clinical} edits (Section~\ref{sec:taxonomy}) as one of five subtypes and provide the definition for each below:
\begin{enumerate}[topsep=0.1ex,itemsep=0.1ex]
    \item \textit{Factual error correction}: Clinician revises information in the draft that conflicts with clinical details present in the input context.
    \item \textit{Omission correction}: Clinician adds clinically relevant information absent from the draft.
    \item \textit{Condensation}: Clinician removes information that, while not incorrect, is determined to be extraneous and non-essential.
    \item \textit{Specificity adjustment}: Clinician rewrites lay or imprecise language (e.g., a patient's own description of their symptoms) into standardized clinical terminology, or replaces abstract information with specific clinical details (e.g., ``antibiotics'' $\rightarrow$ ``vancomycin'').
    \item \textit{Confidence-level adjustment}: Clinician perturbs the amount of certainty or hedging.
\end{enumerate}

\subsection{Prevalence of Ungrounded Edits in Real Clinician Edit Data (Section~\ref{sec:taxonomy})}
\label{appendix:ungrounded-prevalence}

Following our definition of \textit{grounded} vs.~\textit{ungrounded} edits in Section~\ref{sec:taxonomy}, we estimate that more than 45\% of clinical notes from our dataset of 30,000 de-identified and anonymized encounters contain one or more ungrounded edits. We estimate this prevalence through a two-stage process. First, we use our LLM-based edit tagger (Section~\ref{sec:edit-tagger}) on the anonymized corpus of 30,000 encounters. 57\% of the notes are flagged by the tagger as containing ungrounded edits. Following manual review of a random subset of 20 de-identified and anonymized encounters by one of the authors with expertise in assessing clinical documentation, we find that 80\% are true positive cases, which leads to an estimated prevalence of $0.57 \times 0.80 \times 100 = 45.6\%$. 

\renewcommand\thetable{B\arabic{table}}
\renewcommand\thefigure{B\arabic{figure}}
\setcounter{table}{0}
\setcounter{figure}{0}
\section{Additional Details on \dataset{}}
\label{appendix:dataset}
Here, we provide additional details on how we construct \dataset{} (Section~\ref{sec:dataset}). 

\subsection{MIMIC-III Cohort Selection}
\label{appendix:mimic-cohort-selection}

As discussed in Section~\ref{sec:dataset}, we follow the cohort selection criteria proposed by \citet{verifact} in the ``Details Involving the Creation of VeriFact-BHC'' section of their Supplementary Appendix, which we replicate below. 
We extract all hospital admissions in the MIMIC-III Clinical Database v1.4~\citep{mimic-iii} belonging to patients that satisfy the following conditions:
\begin{enumerate}[topsep=0.1ex,itemsep=0.1ex]
    \item Patients whose most recent hospital admission contains a discharge summary with Brief Hospital Course (BHC) section extractable by regular-expression pattern matching;
    \item Patients had at least two additional physician notes in the patient's EHR;
    \item Patients had at least ten notes preceding the discharge summary, including those from prior hospital admissions and encounters.
\end{enumerate}
While \citet{verifact} randomly subsample 100 patients satisfying the above criteria for their analysis, we extract all such patients and construct our \dataset{} dataset using the most recent hospital admissions corresponding to 5,821 patients.

\subsection{Pre-Edit BHC Generation: Handling Admissions with Input Context Exceeding Context Window Sizes}

Among the 5,821 admissions extracted as in Appendix~\ref{appendix:mimic-cohort-selection}, some admissions contain a significantly large number of input prior notes, resulting in cases where the concatenation of all input prior notes does not fit into the context windows of the base models after tokenization. 
As such, for each of the three base models, we pre-tokenize the concatenation of all prior notes available for each admission, and exclude the admissions with inputs that exceed the context window of the model. 
Thus, we only use the remaining admissions that fit into the context window of each model for post-edit BHC generation, and in turn, the training and evaluation of models (as detailed in Section~\ref{sec:experiments}). 
See Table~\ref{tab:edit_pair_stats} for the final number of admissions included for each base model. 

\subsection{Post-Edit BHC Generation: Detailed Description of Step 1 (Clinical Content Alignment)}
\label{appendix:step1-details}

As discussed in Section~\ref{sec:dataset}, we provide GPT-5.5 with the pre-edit (i.e., zero-shot generations) and clinician-written BHCs for each admission and prompt it to add or remove clinical details from the pre-edit BHC to match the content in the clinician-written BHC, while preserving the overall documentation style (e.g., paragraph structure, sentence order, prose style). 
In alignment with our taxonomy (Section~\ref{sec:taxonomy}), each edit operation takes one of the following forms: 
factual error correction, omission correction, condensation, specificity adjustment, confidence level adjustment, and retainment. 
If a clinical detail (a discrete piece of clinical information, such as a single medication with its dosage or a single diagnosis with its supporting rationale) is \textit{consistent} between the two BHCs, we either keep the original detail (\textit{retainment}) or update it to match the specificity of the corresponding detail in the clinician-written BHC (e.g., ``prescribed antibiotics'' $\rightarrow$ ``prescribed vancomycin'') (\textit{specificity adjustment} or \textit{confidence adjustment}). 
If a clinical detail is \textit{inconsistent} between the two BHCs, we either remove the original detail or substitute it with correct information to match what the clinician-written BHC states (\textit{factual error correction}). 
If a clinical detail is only discussed in the pre-edit BHC, we remove it if the content is redundant (e.g., restates information covered/implied elsewhere in the pre-edit BHC), not essential to the clinical understanding of a patient's state (e.g., generic statements like ``patient was monitored''), or a verbose narration of a clinical fact that is already conveyed concisely (\textit{condensation}). 
Clinical details only discussed in the clinician-written BHC are always added to the pre-edit BHC (\textit{omission correction}).
Tables~\ref{tab:step_1_edit_ops_gemma_3_4b_it}--\ref{tab:step_1_edit_ops_llama_3_1_8b_instruct} show the distribution of edit operations resulting from \textbf{Step 1} of the post-edit generation process, each corresponding to Gemma-3-4B-IT, Qwen-2.5-7B-Instruct, and Llama-3.1-8B-Instruct, respectively. 

%% gemma-3-4b-it  (n_admissions = 4,936)
  \begin{table*}[t!]
    \centering
    \caption{Frequency of edit operations from \textbf{Step 1} (clinical content alignment) of our edit simulation process (Section~\ref{sec:dataset}) for Gemma-3-4B-IT, across all admissions in \dataset{}. Coverage denotes the \% of admissions in which the operation appears at least once.}
    \label{tab:step_1_edit_ops_gemma_3_4b_it}
    \resizebox{0.7\linewidth}{!}{%
    \begin{tabular}{lrrrr}
    \toprule
     & \multicolumn{2}{c}{Per operation} & \multicolumn{2}{c}{Per admission} \\
    \cmidrule(lr){2-3} \cmidrule(lr){4-5}
    Edit operation              & Count    & \% of total & Mean count & Coverage (\%) \\
    \midrule
    Specificity Adjustment      & 17{,}298 &  23.7       & 3.50       & 97.3 \\
    Omission Correction         & 20{,}455 &  28.1       & 4.14       & 98.2 \\
    Retain                      & 12{,}123 &  16.6       & 2.46       & 92.1 \\
    Factual Error Correction    & 12{,}383 &  17.0       & 2.51       & 93.8 \\
    Condensation                &  9{,}744 &  13.4       & 1.97       & 88.7 \\
    Confidence Level Adjustment &      831 &   1.1       & 0.17       & 16.2 \\
    \midrule
    Total                       & 72{,}834 & 100.0       & 14.76      & ---  \\
    \bottomrule
    \end{tabular}
    }
  \end{table*}

  %% qwen-2.5-7b-instruct  (n_admissions = 3,469)
  \begin{table*}[t!]
    \centering
    \caption{Frequency of edit operations from \textbf{Step 1} (clinical content alignment) of our edit simulation process (Section~\ref{sec:dataset}) for Qwen-2.5-7B-Instruct, across all admissions in \dataset{}. Coverage denotes the \% of admissions in which the operation appears at least once.}
    \label{tab:step_1_edit_ops_qwen_2_5_7b_instruct}
    \resizebox{0.7\linewidth}{!}{%
    \begin{tabular}{lrrrr}
    \toprule
     & \multicolumn{2}{c}{Per operation} & \multicolumn{2}{c}{Per admission} \\
    \cmidrule(lr){2-3} \cmidrule(lr){4-5}
    Edit operation              & Count    & \% of total & Mean count & Coverage (\%) \\
    \midrule
    Specificity Adjustment      & 11{,}698 &  25.0       & 3.37       & 98.2 \\
    Omission Correction         & 14{,}755 &  31.5       & 4.25       & 98.6 \\
    Retain                      &  8{,}758 &  18.7       & 2.52       & 94.7 \\
    Factual Error Correction    &  6{,}890 &  14.7       & 1.99       & 88.9 \\
    Condensation                &  4{,}261 &   9.1       & 1.23       & 69.6 \\
    Confidence Level Adjustment &      446 &   1.0       & 0.13       & 12.3 \\
    \midrule
    Total                       & 46{,}808 & 100.0       & 13.49      & ---  \\
    \bottomrule
    \end{tabular}
    }
  \end{table*}

  %% llama-3.1-8b-instruct  (n_admissions = 5,037)
  \begin{table*}[t!]
    \centering
    \caption{Frequency of edit operations from \textbf{Step 1} (clinical content alignment) of our edit simulation process (Section~\ref{sec:dataset}) for Llama-3.1-8B-Instruct, across all admissions in \dataset{}. Coverage denotes the \% of admissions in which the operation appears at least once.}
    \label{tab:step_1_edit_ops_llama_3_1_8b_instruct}
    \resizebox{0.7\linewidth}{!}{%
    \begin{tabular}{lrrrr}
    \toprule
     & \multicolumn{2}{c}{Per operation} & \multicolumn{2}{c}{Per admission} \\
    \cmidrule(lr){2-3} \cmidrule(lr){4-5}
    Edit operation              & Count    & \% of total & Mean count & Coverage (\%) \\
    \midrule
    Specificity Adjustment      & 20{,}800 &  27.1       & 4.13       & 99.2 \\
    Omission Correction         & 17{,}814 &  23.2       & 3.54       & 96.5 \\
    Retain                      & 14{,}972 &  19.5       & 2.97       & 96.7 \\
    Factual Error Correction    & 12{,}806 &  16.7       & 2.54       & 93.4 \\
    Condensation                &  9{,}421 &  12.3       & 1.87       & 84.0 \\
    Confidence Level Adjustment &      960 &   1.3       & 0.19       & 18.0 \\
    \midrule
    Total                       & 76{,}773 & 100.0       & 15.24      & ---  \\
    \bottomrule
    \end{tabular}
    }
  \end{table*}

\subsection{Post-Processing of Zero-Shot Base-Model Generations}
\label{appendix:postprocessing}

In practice, we observe that the zero-shot BHC generations from the base models may: (i) not be valid BHCs (e.g., do not read as summaries that a clinician would write, regurgitate the input prior notes); (ii) generate unnecessary content (e.g., conversational text like \texttt{``Okay, here is the brief hospital course section for the patient:''} or section headers like \texttt{``**BHC:**''}); or (iii) repeat the same content until consuming the entire output token budget. 
Simulating edits on such generations can result in cases where either the edit itself is not reflective of the types of post-deployment edits that clinicians would make in practice, or the post-edit version reads as a complete rewrite of the original LLM-generated draft. 

To avoid these issues, we always post-process the zero-shot generated draft by applying the following steps:
\begin{enumerate}[topsep=0.5ex,itemsep=0.1ex]
    \item \textbf{Section header removal:} We strip leading section headers (e.g., \texttt{Brief Hospital Course:}, \texttt{**BHC:**}, \texttt{\#\# Brief Hospital Course (BHC)}) as well as inline header prefixes (e.g., \texttt{BHC: <content>}) by matching a set of regular expressions anchored to the start of the text, applied iteratively (up to three passes) to handle stacked headers.
    \item \textbf{EHR-regurgitation removal:} We truncate the output at the earliest line-anchored occurrence of any marker indicating that the model has begun echoing the raw EHR export, such as note-index dividers (e.g., \texttt{\#\# Note X/Y}), long underscore separators, \texttt{------Protected Section------} banners, field labels like \texttt{Admitting Diagnosis:}, or radiology placeholder tokens (e.g., \texttt{[**Clip Number (Radiology) ...**]}).
    \item \textbf{Repetition removal:} 
    We extract the trailing suffix of the output at each of several fixed lengths ($200, 150, 100, 80, 60$ characters), and for each suffix we search for additional exact occurrences earlier in the text; if a given suffix recurs at least twice with gaps between consecutive occurrences that are all within 15\% of the mean gap (signaling a periodic loop rather than coincidental repetition), we record a candidate cut at the end of its first occurrence, and truncate the output at the earliest such cut across all suffix lengths.
    \item \textbf{Truncation removal:} If the remaining output ends without a sentence-ending punctuation, we truncate everything after the last position matching \texttt{[.!?]} (optionally followed by a closing quote or bracket and then whitespace or end-of-text), except when it would remove more than 30\% of the text (heuristic safeguard against cases where the output is not actually truncated).
\end{enumerate}
The post-processed zero-shot BHC generations then become the pre-edit BHCs upon which \textbf{Step 1} and \textbf{Step 2} of our edit simulation pipeline are subsequently applied. 

\subsection{Post-Edit BHC Generation: Prompts}
\label{appendix:dataset_prompts}

Below, we include the prompts that we use for implementing \textbf{Step 1} and \textbf{Step 2} of the post-edit generation process. For both cases, we always generate the output with greedy decoding. 

\begin{promptbox}[Step 1 - Prompt]
\textcolor{blue}{\textbf{System Prompt}} \\
You are a medical expert in discharge summaries, specifically the Brief Hospital Course (BHC) section. You are simulating realistic clinician edits to LLM-generated BHCs. \\

\textcolor{blue}{\textbf{User Prompt Template}} \\
You will be given: \\
1. An LLM-generated BHC (the "pre-edit BHC draft" that a clinician is editing) \\
2. A clinician-written BHC for the same admission (the gold-standard BHC) \\

\textbf{DEFINITION:} A "clinical detail" is a discrete piece of clinical information in the BHC (e.g., a single diagnosis with its supporting rationale, a single medication with its dose and indication, a single procedure with its outcome). Clinical details are the units of comparison between the two BHCs: each detail may be present in one BHC, both BHCs, or neither, and may be expressed with different levels of specificity, certainty, or terminology across the two. A clinical detail can span multiple atomic facts when those facts are tightly coupled and would naturally be edited together (e.g., a medication and its dose). \\

\textbf{YOUR TASK:} Simulate a clinician editing the LLM-generated BHC draft so that its clinical content aligns with the clinician-written BHC. Decompose each BHC into its constituent clinical details, then for each detail in the LLM-generated BHC, cross-reference it against the corresponding detail (if any) in the clinician-written BHC to decide what edit should be made. The resulting post-edit BHC should look like a **minimally edited version of the LLM draft**, not a rewrite. \\

\textbf{EDITING RULES:} \\
- Preserve the LLM draft's overall paragraph structure, sentence order, and prose style. Do NOT reorganize into problem-oriented format, add headers, or convert between prose and bullets unless the LLM draft already uses that format. \\

- For clinical details discussed in both BHCs with *equivalent* content: Keep the LLM's phrasing as-is. Use the action "retain". \\

- For clinical details discussed in both BHCs where the clinician version contains a *specific clinical fact* that the LLM version generalizes or omits (e.g., a specific drug name, dose, route, frequency, lab value, anatomical location, device specification, or precise timing): Replace the vague phrasing with the specific fact, preserving the surrounding sentence structure. Use the action "specificity\_adjustment". Do NOT use this action for paraphrasings or wording variations that don't add a concrete clinical fact. If the LLM and clinician versions convey the same clinical information using different words, use "retain" — wording differences alone are not specificity adjustments. \\

- For clinical details discussed in both BHCs where the clinician version expresses a different *level of clinical certainty* (e.g., the LLM-generated BHC asserts "the patient had sepsis", while the clinician-written BHC says "the patient was treated empirically for possible sepsis"): Adjust the LLM version's hedging to match the clinician-written version. Use the action "confidence\_level\_adjustment". \\

- For clinical details discussed in both BHCs with *conflicting* content: Replace the LLM version with the clinician-written version. Use the action "factual\_error\_correction". \\

- For clinical details only discussed in the LLM-generated BHC: Decide whether to remove or retain based on the *type* of content. Use the action "condensation" ONLY when the LLM-only content is one of the following: (a) redundant content that restates information already covered elsewhere in the LLM draft; (b) low-value procedural or process detail (e.g., generic statements like "the patient was monitored", "the patient tolerated the procedure well") that doesn't advance clinical understanding; or (c) verbose narration of a fact already conveyed more concisely. For LLM-only content that does NOT fit the above (e.g., a specific clinical finding, a meaningful diagnostic or therapeutic step, a notable event during the admission): Use the action "retain". Many LLM-only details are clinically valid — clinicians vary in what they choose to include in a BHC, and not every omission by the clinician means the content should be removed. \\

- For clinical details only discussed in the clinician-written BHC: Add them to the LLM draft where they fit naturally without breaking coherence or contradicting existing content. Keep the additions concise and consistent with the LLM's writing style. Use the action "omission\_correction". \\

- The final post-edit note should be approximately similar in length to the LLM draft. Small increases are fine for necessary additions, but avoid overinflating the length. \\

\textbf{PROCESS:} First, produce an edit plan listing each change (or explicit retain) from start to end of the document. Then, produce the simulated post-edit note. \\

\textbf{LLM-Generated BHC (Pre-Edit Draft)} \\
\{llm\_bhc\} \\

\textbf{Clinician-Written BHC (Gold-Standard BHC)} \\
\{human\_bhc\} \\
\end{promptbox}

\begin{promptbox}[Step 2 - Prompt]
\textcolor{blue}{\textbf{System Prompt}} \\
You are a medical expert who reformats clinical notes to match a clinician's preferred documentation style. \\

\textcolor{blue}{\textbf{User Prompt Template}} \\
You will be given:\\
1. A *clinician-edited* BHC, where an LLM-generated BHC was edited by a clinician to ensure that the clinical content is correct\\
2. A *clinician-written* BHC showing the clinician's preferred documentation style\\

\textbf{YOUR TASK:} Reformat the edited BHC to match the clinician-written BHC's structural and stylistic conventions. Do NOT change any clinical content; only change formatting, organization, and writing style.\\

\textbf{Aspects to align:} \\
- Section headers / problem-oriented structure vs.~free-form prose\\
- Bullet points vs.~flowing paragraphs\\
- Abbreviation style (e.g., "year old" vs "y.~o.~", "status post" vs "s/p", "follow-up" vs "f/u")\\
- Sentence fragments vs. complete sentences\\
- Problem ordering and grouping\\

Produce ONLY the reformatted note. Do not add or remove clinical content. \\

\textbf{Edited BHC (Content-Correct, Needs Reformatting)} \\
\{step\_1\_output\} \\

\textbf{Clinician-Written BHC (Reflects Preferred Documentation Style)}\\
\{human\_bhc\}\\

\textbf{Reformatted Note}\\
\end{promptbox}

\subsection{Final Dataset Statistics}
\label{appendix:data-statistics}

\begin{table*}[!h]
  \centering
  \caption{Statistics for the model-specific edit-pair datasets. For each base model, we report sample counts and BHC length distributions for the train and validation sets for their corresponding edit pairs.
  Pre-edit BHCs are the zero-shot generations from each model (after post-processing); post-edit BHCs are the final outputs of our two-stage edit simulation pipeline (Section~\ref{sec:dataset}). 
  }
  \label{tab:edit_pair_stats}
  \resizebox{\linewidth}{!}{%
  \begin{tabular}{llrcccc}
  \toprule
   & & & \multicolumn{2}{c}{Pre-Edit BHC (words)} & \multicolumn{2}{c}{Post-Edit BHC (words)} \\
  \cmidrule(lr){4-5} \cmidrule(lr){6-7}
  Base Model & Split & $N$ & Mean (SD) & Median & Mean (SD) & Median \\
  \midrule
  \multirow{2}{*}{Gemma-3-4B-IT}
    & Train       & 4{,}680 & 191.6 \,(98.7)  & 180 & 368.8 \,(162.6) & 346 \\
    & Validation         &    256  & 184.1 \,(58.6)  & 174 & 354.1 \,(161.1) & 334 \\
  \midrule
  \multirow{2}{*}{Qwen-2.5-7B-Instruct}
    & Train       & 3{,}302 & 107.0 \,(42.3)  &  98 & 305.3 \,(138.1) & 282 \\
    & Validation         &    167  & 104.6 \,(40.8)  &  96 & 290.8 \,(139.0) & 267 \\
  \midrule
  \multirow{2}{*}{Llama-3.1-8B-Instruct}
    & Train       & 4{,}775 & 252.4 \,(142.9) & 229 & 378.5 \,(170.4) & 353 \\
    & Validation         &    262  & 254.0 \,(106.2) & 244 & 381.2 \,(184.8) & 350 \\
  \bottomrule
  \end{tabular}
  }
\end{table*}

The final summary statistics for our edit pair dataset are shown in Table~\ref{tab:edit_pair_stats}. 
The edit datasets for each model follow from: (i) filtering out the admissions whose input context exceeds the context window size of the given model; (ii) zero-shot generating the BHCs; (iii) applying the post-processing pipeline (Appendix~\ref{appendix:postprocessing}); (iv) running \textbf{Step 1}; and finally, (v) running \textbf{Step 2}. 
We focus on generating the post-edit BHCs for the train and validation sets of \dataset{}, since we only need the clinician-written BHCs as references for evaluation on the test sets. 

\subsection{Details and Annotation Guide for Expert Validation of Simulated Edits}
\label{appendix:edit-annotation}

Below, we include the definitions of ratings for our expert annotation tasks described in Section~\ref{sec:edit-annotation}:

\textbf{Task 1: Coherence \& Real Documentation Practice. }
\begin{itemize}[topsep=0.1ex,itemsep=0.1ex]
    \item \textbf{Reasonable}: Resulting text is coherent with other parts of the post-edit draft and written in professional language, organization, and style appropriate for clinical documentation.
    \item \textbf{Unreasonable}: Resulting text has at least one clear issue in terms of coherence with other parts of the post-edit draft or the language, organization, or style used.
    \item \textbf{Unsure}: Unclear whether the coherence and language used are reasonable or unreasonable. 
\end{itemize}

\textbf{Task 2: Clinical Content Alignment to the Reference.}
\begin{itemize}[topsep=0.1ex,itemsep=0.1ex]
    \item \textbf{Aligned}: The resulting changes only (i) substitute/add in clinical details substantiated by the reference BHC and (ii) remove clinical details that are not consistent with the reference BHC.
    \item \textbf{Not Aligned}: The resulting changes either (i) substitute/add in clinical details that are not substantiated by the reference BHC or (ii) remove clinical details that are consistent with the reference BHC and integral to the summarization.
    \item \textbf{Unsure}: It is unclear whether the given edit is aligned or not. \item \textbf{N/A}: The edit does not substantively modify the clinical content present in the BHC. 
\end{itemize}

We also include the full annotation guide provided to the NP annotators:

\begin{promptbox}[Annotation Guide]

We are evaluating the quality of edits made to an LLM-generated discharge summary, specifically to its brief hospital course (BHC) section. The BHC section details key clinical events, diagnoses, and treatments for a patient during their hospitalization. The pre-edit draft was obtained by providing an LLM with all of the EHR notes available for a patient prior to their discharge from the hospital. For each hospitalization, there is also a clinician-written reference BHC, which serves as a source of ground-truth. \\

The goal of this annotation task is to check if each individual edit to the BHC draft is “reasonable”, considering: \\
- Whether the resulting changes are coherent with other parts of the final BHC draft and written in professional language, organization, and style appropriate for clinical documentation.\\
- Whether they only (i) substitute/add in clinical details substantiated by the reference BHC and (ii) remove clinical details that are not consistent with the reference BHC.\\

\textcolor{blue}{\textbf{What You Will See}}\\
Each annotation example will include the following items:\\
- A clinician-written reference BHC\\
- An LLM-generated draft of the BHC (pre-edit BHC)\\
- The final BHC draft (post-edit BHC)\\
- One of the edits made to the pre-edit BHC\\

In the pre-edit and post-edit BHCs, the text added/removed via the edit are also demarcated with “\{\{ \}\}”.\\

\textcolor{blue}{\textbf{What To Do}}\\
For each edit, determine whether it reads as a “reasonable” edit, considering the description above. Specifically, provide the following information:\\

1-1.~Coherence \& Documentation Practice\\
- Reasonable: The resulting text is coherent with other parts of the post-edit draft and written in professional language, organization, and style appropriate for clinical documentation.\\
- Unreasonable: The resulting changes have at least one clear issue in terms of coherence with other parts of the post-edit draft or the language, organization, or style used.\\
- Unsure: You are unsure whether the coherence and language used are reasonable or unreasonable.\\

1-2.~Free-Text Comment \\
Please briefly explain your reasoning.\\

2-1.~Clinical Content Alignment to Reference\\
- Aligned: The resulting changes only (i) substitute/add in clinical details substantiated by the reference BHC and (ii) remove clinical details that are not consistent with the reference BHC.\\
- Not Aligned: The resulting changes either (i) substitute/add in clinical details that are not substantiated by the reference BHC or (ii) remove clinical details that are consistent with the reference BHC and integral to the summarization.\\
- Unsure: You are unsure whether the given edit is aligned or not.\\
- N/A: The edit does not substantively modify the clinical content present in the BHC.\\

2-2.~Free-Text Comment\\
Please briefly explain your reasoning.\\

\textcolor{blue}{\textbf{Important Notes (Please Read)}}\\
- “Reasonable” does not mean “you would personally make this change”. The focus should be on whether an edit clearly degrades the clinical quality or utility of the BHC. \\
- Because all BHCs have undergone a de-identification process, the pre-/post-edit BHCs may contain redactions (e.g., “\_\_\_\_\_”, “**First Name4**”, “**Hospital1**”, “NamePattern2”) or imputed dates (e.g., “[**2122-6-20**]”) that render the BHCs awkward or grammatically incorrect. Please ignore such de-identification artifacts when determining the “reasonability” of edits.

\end{promptbox}

\renewcommand\thetable{C\arabic{table}}
\renewcommand\thefigure{C\arabic{figure}}
\setcounter{table}{0}
\setcounter{figure}{0}
\section{Additional Details on the Edit Tagger}
\label{appendix:edit-tagger}
\subsection{Prompt for Tagging Edits}
\label{appendix:tagging_prompt}

\begin{promptbox}[Edit Tagging]

\textcolor{blue}{\textbf{System Prompt}} \\
You are a medical expert reviewing clinician edits made to AI-generated discharge summaries (in particular, the Brief Hospital Course (BHC) section).
The discharge summary generation model receives the patient's pre-discharge EHR notes as input, and produces a draft discharge summary. A clinician then reviews the draft and makes edits to produce the final discharge summary.
While clinicians generally make edits to improve upon the original AI-generated BHC, keep in mind that they may make edits that introduce new errors, add information unsubstantiated by the pre-discharge notes, or serve purely stylistic purposes. \\

\textcolor{blue}{\textbf{User Prompt Template}}

You will be given: \\
1.~An LLM-generated BHC draft (the "draft" that a clinician is editing) \\
2.~A final clinician-edited BHC for the same admission \\ 
3.~All pre-discharge EHR notes available for the patient \\

\textbf{YOUR TASK:} 
Identify all observable atomic edits between the LLM-generated BHC draft and the final edited BHC, then characterize each edit. \\

\textbf{DEFINITION:} \\
- An atomic edit is the smallest unit of change between the draft and final BHC that corresponds to a single reason for modification. It should correspond to one specific reason why the clinician modified the BHC. All text changes driven by that reason belong to the same edit; changes driven by different reasons should be considered as separate edits. For example, if a clinician changes "the patient had a fever and cough" to "the patient had a fever", this would be one atomic edit corresponding to the reason of correcting an unsupported symptom (cough). \\
- Split edits if they differ in: \\
(a) type of content being edited: clinical vs non-clinical \\
(b) grounding (supported vs. unsupported by EHR) \\
(c) the clinician's intent (e.g., correcting a factual error vs. correcting an omission vs. making a stylistic change) \\

For each edit, provide the following information: \\
(1) a concise description of the textual changes resulting from the atomic edit \\
(1a) the surface edit operation: addition, substitution, or deletion \\
(1b) the exact verbatim text span from the Original BHC corresponding to this atomic edit \\
- copy the text exactly as it appears in the Original BHC \\
- if the atomic edit is an addition with no relevant Original-BHC span, return an empty string \\
- prefer the smallest contiguous text span that uniquely identifies the edit \\
(1c) the exact verbatim text span from the Final Edited BHC corresponding to this atomic edit \\
- copy the text exactly as it appears in the Final Edited BHC \\
- if the atomic edit is a deletion with no relevant Final-Edited-BHC span, return an empty string \\
- prefer the smallest contiguous text span that uniquely identifies the edit \\

(2) whether the edit modifies clinical content in the LLM-generated BHC \\
- clinical edits are edits that modify the clinical content of the BHC, such as changes to diagnoses, treatments, symptoms, or other medical information. \\
- non-clinical edits are edits that leave the clinical content unchanged and modifies formatting, organization, grammar fixes, or stylistic properties. \\

(3) if the edit is a clinical edit, determine whether the edit is clearly supported by the pre-discharge EHR notes \\
- grounded: Supported by EHR notes (e.g., a diagnosis mentioned in the BHC is also mentioned in the EHR notes) \\
- ungrounded: Not supported by EHR notes (e.g., a diagnosis mentioned in the BHC is not mentioned in any of the EHR notes) \\

(4) reasoning for whether the edit is or isn't substantiated by the pre-discharge EHR notes \\

(5) if the edit is ungrounded, select one of the following categories that best describes the edit's relationship to the pre-discharge EHR notes: \\
- contradictory: conflicts with information in the pre-discharge notes \\
- unsupported: not explicitly present and not inferable from EHR \\
- inferred but uncertain: not explicitly stated but could be inferred from the available information, but there is doubt as to whether or not the inference is correct \\

(6) if the edit is a clinical edit, select one of the following categories that best describes the edit type: \\
- factual\_error\_correction: fixes incorrect or unsupported content in draft \\
- omission\_correction: adds missing clinically relevant information supported by EHR \\
- refinement: modifies specificity, detail (e.g., adding more specific terms or taking out details), or certainty of clinical content without correcting a factual error or adding omitted content\\
- other (if the edit does not fit into any of the above categories)\\

(7) if the edit is a clinical refinement edit, select one of the following categories that best describes the edit type:\\
- condensation of content: when the clinician takes out information that is not conflicting with the EHR data\\
- specificity adjustment: when the clinician adds more specific information that is supported by the EHR data (e.g., changing "antibiotics" to "vancomycin and cefepime")\\
- hedging/certainty change: when the clinician changes the certainty level of a statement (e.g., changing "the patient had a fever" to "the patient may have had a fever")\\

(8) if the edit is a nonclinical edit, select one of the following categories that best describes the edit type:\\
- formatting: changes related to the structure, layout, or organization of the text (e.g., adding bullet points, changing section headers)\\
- grammar: changes that fix grammatical errors, punctuation, or spelling\\
- wording: changes that adjust phrasing or terminology without altering clinical content\\
- other: any other non-clinical edits that do not fit into the above categories\\

(9) a description of the clinician's intent underlying the edit, e.g.:\\
- Example 1: The clinician corrected an imprecise admission diagnosis in the AI-generated note by substituting in the specific clinical diagnosis established during the hospitalization.\\
- Example 2: The clinician corrected an incomplete description of the patient's antibiotic course by substituting in the specific agents used and the de-escalation that occurred during the hospitalization.\\
- Example 3: The clinician added a consultation and its recommendations that occurred during the hospitalization but were omitted from the AI-generated note.\\
- Example 4: The clinician removed speculative differential diagnosis content that is not appropriate for the summary nature of the brief hospital course.\\

(10) based on the clinician's intent and the nature of the edit, your judgment if this is a global vs. personal edit:\\
- global edits are edits that you think most clinicians would make given the same LLM-generated BHC draft and the same pre-discharge EHR notes. These edits are driven by reasons that are likely to be shared across different clinicians.\\
- personal edits are edits that may be driven by individual clinician preferences, such as stylistic choices or different thresholds for including certain information in the BHC. These edits are less likely to be shared across different clinicians.\\

\textbf{IMPORTANT:}\\
- Do not merge changes driven by different reasons into a single edit.\\
- The `pre\_span\_text` and `post\_span\_text` fields must be copied verbatim from the corresponding BHC text, not paraphrased.\\
- If incorrect clinical content is replaced with correct clinical content in the edit, this should be treated as correcting a factual error, not an omission of the added content. In this sense, factual-error corrections and omission corrections are mutually exclusive for a given atomic edit.\\
- For each item, keep your response concise and clear, around 1-2 sentences.\\

\vspace{1em}

\end{promptbox}

\subsection{Edit Distribution}
\label{appendix:edit-distribution}

We report the distribution of edits in our \textbf{Step 1} and \textbf{Step 2} post-edit BHC generations in Table \ref{tab:edit-distribution}.

\begin{table*}[h!]
\centering
\caption{Distribution of edit types across models and admissions included. 
\textbf{Step 1} and \textbf{Step 2} show the distributions after each step of the post-edit simulation process (Section~\ref{sec:dataset}), with atomic edits extracted with respect to the pre-edit BHCs. 
``N/A'' groundedness corresponds to non-clinical edits, since it is only assessed for clinical edits. 
Mean denotes the average number of edits per note.}
\label{tab:edit-distribution}
\small
\setlength{\tabcolsep}{4pt}
\begin{adjustbox}{width=\textwidth}
\begin{tabular}{llrrcccccccccc}
\toprule
& & \multicolumn{2}{c}{Count} & \multicolumn{3}{c}{Operation (\%)} & \multicolumn{2}{c}{Clinical (\%)} & \multicolumn{3}{c}{Groundedness (\%)} & \multicolumn{2}{c}{Scope (\%)} \\
\cmidrule(lr){3-4}\cmidrule(lr){5-7}\cmidrule(lr){8-9}\cmidrule(lr){10-12}\cmidrule(lr){13-14}
Model & Setting & Total & Mean & Add & Sub & Del. & Clin. & Non-C. & Grnd. & Ungr. & N/A & Global & Pers. \\
\midrule
\multirow{2}{*}{Gemma-3-4B-IT} & \textbf{Step 1} & 99{,}960  & 20.25 & 53.15 & 28.28 & 18.57 & 97.17 & 2.83  & 64.97 & 30.93 & 2.83  & 76.65 & 23.35 \\
                       & \textbf{Step 2} & 122{,}516 & 24.82 & 55.71 & 27.31 & 16.99 & 89.15 & 10.85 & 61.30 & 26.34 & 10.85 & 70.73 & 29.27 \\
\midrule
\multirow{2}{*}{Qwen-2.5-7B-Instruct}  & \textbf{Step 1} & 64{,}377  & 18.56 & 59.58 & 27.85 & 12.57 & 98.59 & 1.41  & 61.27 & 36.14 & 1.41  & 73.41 & 26.59 \\
                       & \textbf{Step 2} & 80{,}455  & 23.19 & 60.22 & 28.66 & 11.13 & 89.42 & 10.58 & 57.06 & 31.05 & 10.58 & 67.90 & 32.10 \\
\midrule
\multirow{2}{*}{Llama-3.1-8B-Instruct} & \textbf{Step 1} & 103{,}812 & 20.61 & 51.34 & 32.73 & 15.94 & 98.14 & 1.86  & 65.12 & 31.89 & 1.86  & 74.59 & 25.41 \\
                       & \textbf{Step 2} & 126{,}503 & 25.11 & 53.37 & 31.60 & 15.02 & 88.72 & 11.28 & 60.28 & 27.09 & 11.28 & 67.25 & 32.75 \\
\bottomrule
\end{tabular}
\end{adjustbox}
\end{table*}

\subsection{Details on Edit Tagger Evaluation}
\label{appendix:tagger-eval}

\paragraph{Clinical vs.~Non-Clinical \& Global vs.~Personal Edits.} As discussed in Section~\ref{sec:edit-tagger}, we assess the edit tagger judgments along the \textit{Clinical vs.~Non-Clinical} and \textit{Global vs.~Personal} axes by collecting expert annotations from nurse practitioners (NPs) with hospitalist backgrounds. 
In order to achieve uniform coverage over different base LLMs---i.e., Gemma-3-4B-IT, Qwen-2.5-7B-Instruct, and Llama-3.1-8B-Instruct---and all types of edits (as determined by the edit tagger), we use a stratified sampling approach for selecting the 25 atomic edits to annotate, which all come from the training and validation sets of \dataset{}. 
Given that we annotate each atomic edit along both \textit{Clinical vs.~Non-Clinical} and \textit{Global vs.~Personal} axes simultaneously, we first allocate roughly the same number of examples to sample from each axis combination: 7 clinical \& global edits, 6 clinical \& personal edits, 6 non-clinical \& global edits, and 6 non-clinical \& personal edits. 
Then, for each combination, we sample roughly the same number of edits from each base LLM. 
Through this process, we obtain an annotation set that consists of 13 clinical vs.~12 non-clinical edits, 13 global vs.~12 personal edits, with 9 edits from Gemma-3-4B-IT, 8 edits from Qwen-2.5-7B-Instruct, and 8 edits from Llama-3.1-8B-Instruct. 
We note that our sampling procedure upsamples certain combinations of the two axes that occur less frequently in practice (e.g., non-clinical edits are 89\% personal in \dataset{}, according to the edit tagger outputs), and thus our annotation set may not be reflective of the true distribution of edit types in \dataset{}. 

Below, we include the full annotation guide provided to the NP annotators:

\begin{promptbox}[Annotation Guide]

We are evaluating the accuracy of an edit tagger, which extracts and categorizes all atomic edits observable between the pre-edit and post-edit versions of a discharge summary. We specifically focus on the brief hospital course (BHC) section of each discharge summary, which details the key clinical events, diagnoses, and treatments for a patient. The pre-edit draft was obtained by providing an LLM with all of the EHR notes available for a patient prior to their discharge from the hospital. \\

The edit tagger categorizes each edit along each the following axes:\\

\textbf{1. Clinical vs.~Non-Clinical} \\
- Clinical edits are changes that alter the clinical content in the original draft (e.g., factual error correction, omission correction, condensation, specificity adjustment, and confidence-level adjustment), impacting the characterization of the patient’s past and current clinical states. For instance, specificity adjustment such as “recent headache” → “headache since 5 days ago” or factual error correction such as “He was transferred to the CICU” → “He was transferred to the SICU” are examples of Clinical edits that change the clinical characterization of patients.\\
- Non-clinical edits are changes that leave clinical content intact and instead only modify the formatting, organization, or other stylistic properties of the BHC. For instance, a change from free-form prose to a bullet-pointed list of problems or a change from contracted abbreviations to expanded medical terms would be examples of Non-Clinical edits, assuming that they do not meaningfully change the documented clinical content.\\

\textbf{2. Global vs.~Personal} \\
- Global edits are changes that the general population of hospitalists/practitioners would make to improve the clinical accuracy and quality of the BHC. For instance, an edit that alters core clinical details directly relevant to the key problems and events during a patient’s hospitalization is likely a correction that most hospitalists/practitioners would make. \\
- Personal edits are changes that reflect local preferences, which can vary across hospitalists/practitioners. For instance, when discussing the same problem exhibited by a patient, one hospitalist/practitioner may prefer to provide extensive context including details that are not integral to its discussion, while another hospitalist/practitioner may prefer to document in a focused manner. As another example, a hospitalist/practitioner may prefer to use different terms or language to describe the same clinical content (e.g., hedging language).\\

The goal of this annotation task is to annotate whether a given edit is Clinical vs. Non-Clinical and Global vs. Personal, which will serve as labels for validating the edit tagger outputs.\\

\textcolor{blue}{\textbf{What You Will See}}\\
Each annotation example will include the following:\\
- An LLM-generated draft of a BHC section (pre-edit BHC)\\
- The final BHC draft (post-edit BHC)\\
- One of the edits made to the pre-edit BHC\\

In the pre-edit and post-edit BHCs, the text added/removed via the edit are also demarcated with “\{\{ \}\}”.\\

\textcolor{blue}{\textbf{What To Do}}\\
For each annotation example, please provide the following information:\\

1-1.~Clinical vs.~Non-Clinical Label\\
- Clinical: The edit corresponds to a change that alters the clinical content in the original draft (e.g., factual error correction, omission correction, condensation, specificity adjustment, and confidence-level adjustment), impacting the characterization of the patient’s past and current clinical states.\\
- Non-Clinical: The edit corresponds to a change that leaves clinical content intact and instead only modifies the formatting, organization, or other stylistic properties of the BHC.\\
- Unsure: You should select this label if you find the choice ambiguous.\\

1-2.~Free-Text Comment\\
Please briefly explain your reasoning for the Clinical vs. Non-Clinical annotation task.\\

2-1.~Global vs.~Personal Label\\
- Global: The edit corresponds to a change that hospitalists/practitioners would generally make to improve the clinical accuracy and quality of the draft.\\
- Personal: The edit reflects local preferences that vary across hospitalists/practitioners.\\
- Unsure: You should select this label if you find the choice ambiguous.\\

2-2.~Free-Text Comment\\
Please briefly explain your reasoning for the Global vs. Personal annotation task.\\

\textcolor{blue}{\textbf{Important Notes (Please Read)}}\\
- Because all BHCs have undergone a de-identification process, the pre-/post-edit BHCs may contain redactions (e.g., “\_\_\_\_\_”, “**First Name4**”, “**Hospital1**”, “NamePattern2”) or imputed dates (e.g., “[**2122-6-20**]”) that render the BHCs awkward or grammatically incorrect. Please ignore such de-identification artifacts when annotating the edits.

\end{promptbox}

\paragraph{Grounded vs.~Ungrounded Edits.} We evaluate the edit tagger's groundedness judgments against \verifactbhc{} \citep{verifact}, which includes clinician annotations of whether the claims in the BHCs \citet{verifact} review are supported by the prior EHR notes. 
To do so, we zero-shot generate the pre-edit BHCs from the three base models (Gemma-3-4B-IT, Qwen-2.5-7B-Instruct, and Llama-3.1-8B-Instruct) and run the post-edit simulation process outlined in Section~\ref{sec:dataset}, for the MIMIC-III admissions annotated in \verifactbhc{}. 
Note that the resulting edit pairs are only used for evaluating the edit tagger's groundedness judgments and never used for model fine-tuning, 
as these admissions are held out for constructing the \textit{grounded} test set (Section~\ref{sec:experiments}). 
Since \citet{verifact} annotate groundedness at a different granularity, there is no exact one-to-one correspondence between the units they annotate and the units our edit tagger produces (atomic edits). 
We therefore construct a comparable subset as follows: we restrict to clinical edits whose text spans align to at least one annotated proposition (via lexical similarity to \textsc{VeriFact-BHC} chunks), then collapse the matched proposition verdicts into a single edit-level label---treating \textit{Supported} as grounded and \textit{Not Supported} or \textit{Not Addressed} as ungrounded. 
Edits that match no propositions are excluded from evaluation. To ensure the validity of our evaluation, we conduct manual review of this procedure to mainly verify that text spans are aligned reasonably and resolve edits whose matched propositions yield mixed verdicts. We conduct manual evaluation on a randomly sampled ($n=150$) subset per model. Note that since each model has different edits, this does not necessarily cover the same set of admissions for each model.
On the subsets for each model, our tagger achieves an overall agreement rate of 78\%, 80\%, and 75.3\% with the clinician annotations for Gemma-3-4B-IT, Qwen-2.5-7B-Instruct, and Llama-3.1-8B-Instruct, respectively. Experts agree unanimously with each other on 63.1\% (149/236), 64.5\% (156/242), and 63.9\% (138/216) of cases, respectively. Note that although 150 atomic edits were sampled per model for analysis, each edit could align with multiple VeriFact propositions (the unit which was annotated by clinicians), yielding 236, 242, and 216 unique propositions for Gemma-3-4B-IT, Qwen-2.5-7B-Instruct, and Llama-3.1-8B-Instruct, respectively.

\renewcommand\thetable{D\arabic{table}}
\renewcommand\thefigure{D\arabic{figure}}
\setcounter{table}{0}
\setcounter{figure}{0}
\section{Additional Details on Edit Filtering}
\label{appendix:filtering}

\subsection{Details on Edit Filtering Process}
\label{appendix:edit-filtering-details}

For each pair and filter, after we produce a target by reverting every atomic edit whose tags are excluded by the filter, as illustrated in Section \ref{sec:filtering}, 
we apply a ``smoothing'' step based on GPT-5.5 that repairs local fluency artifacts (e.g., broken sentences, dangling references, inconsistent tense). 
To minimize unwanted content changes from smoothing, we implement it with explicit instructions not to alter the propositional content of any retained edit (see prompt in Appendix~\ref{appendix:edit-filtering-prompt}).

\begin{table*}[h]
\centering
\small
\caption{The impact of different smoothing step implementations in edit filtering on SFT results, in the \textit{Grounded}-filter setting. We show the results for Qwen-2.5-7B-Instruct.}
\label{tab:smoothing}
\resizebox{\textwidth}{!}{%
\begin{tabular}{lcccccc}
\toprule
& \multicolumn{3}{c}{Alignment to Clinician-Written Reference} & \multicolumn{3}{c}{Groundedness} \\
\cmidrule(lr){2-4} \cmidrule(lr){5-7}
Setting & DocLens-P & DocLens-R & MEDCON & MedScore & DocLens-Attr. & PDSQI-9 \\
\midrule
Base                                & 0.271 & 0.174 & 0.253 & 0.948 & 0.900 & 0.599 \\
\midrule
SFT (\textbf{Step 2})                           & 0.303 & 0.331 & 0.380 & 0.874 & 0.753 & 0.236 \\
\midrule
SFT (\textit{Grounded}, Default Smoothing)   & 0.320 & 0.325 & 0.371 & 0.953 & 0.880 & 0.580 \\
SFT (\textit{Grounded}, Minimal Smoothing)   & 0.304 & 0.310 & 0.368 & 0.948 & 0.866 & 0.577 \\
\textbf{SFT (\textit{Grounded}, No Smoothing)}        & \textbf{0.334} & \textbf{0.339} & \textbf{0.383} & \textbf{0.954} & \textbf{0.861} & \textbf{0.574} \\
\bottomrule
\end{tabular}
}
\end{table*}

\paragraph{Smoothing ablations.} Here, we show that our findings in Section~\ref{sec:results} are not dependent on having the smoothing step within the edit filtering process. 
To demonstrate this point, we conduct additional ablations with Qwen-2.5-7B-Instruct, where we compare between three settings:
\begin{enumerate}[topsep=0.1ex,itemsep=0.1ex]
    \item \textbf{Default Smoothing:} The default approach based on the prompt in Appendix~\ref{appendix:edit-filtering-prompt}.
    \item \textbf{Minimal Smoothing:} We update the LLM prompt for smoothing to explicitly restrict any repairs to be at the level of punctuation, spacing, duplicated conjunctions, dangling fragments, or sentence boundaries that are directly broken by the edit reversal.
    \item \textbf{No Smoothing:} Edits are reverted with regex, and no repairs are made to fix broken or incoherent sentences. 
\end{enumerate}
In Table~\ref{tab:smoothing}, we see that the choice of smoothing has little impact on the observed trends---i.e., improvements in alignment to reference, less degradation in groundedness compared to when fine-tuning on the raw post-edit BHCs (\textbf{Step 2}). 
Notably, even with no LLM-based smoothing in the filtering step (No Smoothing, in bold), we observe that filtering mitigates degradation in groundedness after SFT, while still allowing improvements over the base model. 

\subsection{Prompt for Filtering Post-Edit BHCs}
\label{appendix:edit-filtering-prompt}

\begin{promptbox}[Filtering Prompt.]
\textcolor{blue}{\textbf{System Prompt}} \\
You are a medical expert revising a clinician-edited brief hospital course (BHC). Some
  edits in the clinician-edited BHC have already been identified as edits that should be reversed under a predefined filter policy. Your job is to remove or revert only those specific edits, while preserving all
  remaining content. \\

  You are not deciding which edits should be reversed. That decision has already been made for
  you. You must execute the requested changes and keep the note coherent. \\

  \textbf{Rules:} \\
  - Apply only the listed edit reversals. \\
  - Do not remove content beyond the listed edits. \\
  - Do not add any new clinical facts. \\
  - Keep changes minimal and local. \\
  - It is allowed to make small wording, punctuation, or sentence-flow repairs that are
  necessary after removing/reverting the listed content. \\
  - Preserve the overall structure and style of the clinician-edited BHC as much as
  possible. \\
  - Output only the final filtered BHC text.\\

\textcolor{blue}{\textbf{User Prompt Template}}\\
You will be given: \\
  1. The original AI-generated BHC draft \\
  2. The clinician-edited BHC \\
  3. A list of already-identified edits that must be reversed \\

  How to interpret the listed edits: \\
  - If the edit operation is `addition`, remove the listed `post\_span\_text` content. \\
  - If the edit operation is `substitution`, replace the listed `post\_span\_text` with the
  listed `pre\_span\_text`. \\
  - If the edit operation is `deletion`, restore the listed `pre\_span\_text` content into
  the note where it naturally belongs. \\

  You may make only small local wording or punctuation repairs that are necessary after
  applying those changes.
  Do not add new clinical facts. Do not remove anything beyond what is needed to reverse
  the listed edits. \\

  \textbf{Original AI-Generated BHC} \\
  \{pre\_edit\_bhc\} \\

  \textbf{Clinician-Edited BHC} \\
  \{post\_edit\_bhc\} \\

  \textbf{Targeted Edits To Reverse} \\
  \{targeted\_edits\} \\
  
\end{promptbox}

\renewcommand\thetable{E\arabic{table}}
\renewcommand\thefigure{E\arabic{figure}}
\setcounter{table}{0}
\setcounter{figure}{0}
\section{Additional Details on Experiments}
\label{appendix:experiments}
\subsection{Model Training} For SFT, we train all models with LoRA adapters for 5 epochs with rank $r=16$, batch size $B=64$, and weight decay $\lambda = 0.01$, using the AdamW optimizer \citep{adamw} with cosine annealing (peak learning rate $\eta = \text{1e-4}$) and a linear warmup ratio of 0.03. 
We add the LoRA adapters to all of the self-attention projection weights and the multi-layer perceptron (MLP) layer weights. 
We use the validation cross-entropy loss for early stopping and use the resulting model for final evaluation on the test set. 

\subsection{Model Inference and Prompts} 
\label{appendix:pre-edit-prompt}

To generate the BHCs from each model (including both the base model and the fine-tuned models), we use the following prompt shown below and sample with greedy decoding. We only include the system prompt for zero-shot generations using the base model. For the SFT runs, we exclude the system prompt in training and evaluation. 
\begin{promptbox}[BHC Generation Prompt]
    \textcolor{blue}{\textbf{System Prompt}}\\
    You are a medical expert on documenting discharge summaries, in particular, its brief hospital course (BHC) section. Given all of the pre-discharge EHR notes available for a patient, your job is to generate the BHC section. The BHC section should be concise and to the point, and all of the clinical details in the BHC section should be clearly substantiated by the pre-discharge notes. Do not output anything other than the BHC section. No headers, explanations, or text corresponding to other sections of the discharge summary.\\

    \textcolor{blue}{\textbf{User Prompt Template}}\\
    Generate the brief hospital course (BHC) section of the discharge summary for this patient. \\

    \#\# Note 1/N: [\{CATEGORY\}] [\{DESCRIPTION\}] [\{CHARTDATE\}] \\
    $\langle$Prior Note$\rangle$\\

    $\vdots$ \\

    \#\# Note N/N: [\{CATEGORY\}] [\{DESCRIPTION\}] [\{CHARTDATE\}] \\
    $\langle$Prior Note$\rangle$\\
    
    **YOUR RESPONSE:**\\
\end{promptbox}

As noted in Section~\ref{appendix:postprocessing}, we find that models may fail to generate a valid BHC or contain degenerate outputs (e.g., repetitions until consuming the full output token budget), especially when zero-shot prompting a base model. 
To mitigate the impact of such artifacts in evaluation, we always apply the same post-processing pipeline to all models. 

\subsection{Construction of the Grounded Test Set (\verifactbhc{})}
\label{appendix:grounded-test-set}

To construct the \textit{grounded} test set based on \verifactbhc{} (Section~\ref{sec:experiments}), where we remove any ungrounded clinical details from the clinician-written reference BHCs, we use the clinician annotations provided by \citet{verifact}. In particular, for each clinician-written reference BHC, we retrieve the atomic claim-level \textit{clinician consensus} annotations on whether the given claim is deemed as ``Supported'' (fully supported by the input prior notes), ``Not Supported'' (partially supported at best), or ``Not Addressed'' (not discussed at all) and remove all claims that are labeled as the latter two. 
To implement this approach, we provide GPT-5.5 with the following prompt below, where ``flagged\_claims\_block'' is a list of atomic claims from \verifactbhc{} with ``Not Supported'' or ``Not Addressed'' labels. 
We emphasize here that the claims removed are solely determined by the clinician annotations from \verifactbhc{} and that GPT-5.5 only carries out the removal. 

\begin{promptbox}[Grounded Test Set - Prompt]
    \textcolor{blue}{\textbf{System Prompt}} \\
    You are a clinical documentation expert. You are removing specified clinical details from a Brief Hospital Course (BHC) section of a discharge summary while preserving the rest of the document verbatim and the document's coherence. \\

    \textcolor{blue}{\textbf{User Prompt Template}} \\
    You will be given:\\
1. A human-written BHC (the document to edit)\\
2. A list of atomic claims that have been flagged for removal, each with a verdict label\\
   - "Not Supported": the claim is addressed in the prior EHR notes but is not fully supported by them\\
   - "Not Addressed": the claim is not addressed at all in the prior EHR notes\\

\textbf{YOUR TASK:} Produce a stripped version of the BHC in which the flagged claims have been removed. The goal is to obtain a version of the BHC that is entirely grounded in the prior EHR notes — every clinical detail in the stripped BHC should correspond to content that is supported by, or at least addressed in, those notes.\\

\textbf{EDITING RULES:}\\
- Preserve all clinical content NOT in the flagged-claims list verbatim. Do not paraphrase, reorganize, or compress supported content.\\
- Preserve the original paragraph structure, section headers (e.g., "\# AMS"), bullets vs. prose, abbreviations, and writing style. The stripped BHC should look like a minimally edited version of the original — not a rewrite.\\
- For each flagged claim, locate the corresponding clinical detail in the BHC and remove it. The flagged claim text is an atomic-fact summary, not a verbatim substring; identify the BHC text that *expresses* the claim and remove that text.\\
- When a flagged claim is fused with supported content in the same sentence (e.g., "The patient was treated with vancomycin for sepsis" where only "for sepsis" is unsupported), remove only the unsupported portion and adjust grammar/punctuation minimally so the remaining sentence reads naturally. Mark these cases as "partially\_removed".\\
- If removing a flagged claim would leave a dangling reference, an empty section, or a grammatically broken sentence, prefer removing the surrounding scaffolding (e.g., the introductory clause or the now-empty header) rather than retaining the unsupported claim. Only mark a claim as "retained" if removing it is genuinely not possible without rewriting supported content.\\
- Section-header cleanup: if removing flagged claims leaves a problem-oriented section header (e.g., "\# DM2", "\# URINARY TRACT INFECTION") with no remaining clinical content beneath it, remove that header as well. Do not leave empty or near-empty sections behind.\\
- Do NOT introduce any new clinical content, hedging language, or transitional phrases. The stripped BHC must be a strict subset of the information in the original BHC.\\
- Do NOT remove content that was not flagged. Supported claims and unflagged content must remain.\\

\textbf{PROCESS:} First, for each flagged claim (in the order given), produce an entry in the edit plan recording how it was handled. Then, produce the stripped BHC.\\

\textbf{Original Human-Written BHC}\\
\{bhc\_text\}\\

\textbf{Flagged Claims}\\
\{flagged\_claims\_block\} \\
\end{promptbox}

\renewcommand\thetable{F\arabic{table}}
\renewcommand\thefigure{F\arabic{figure}}
\setcounter{table}{0}
\setcounter{figure}{0}
\section{Additional Details on Expert Validation of BHC Groundedness}
\label{appendix:groundedness-annotation}
Here, we provide the remaining details for the expert validation study on the groundedness of model-generated BHCs, discussed in Section~\ref{sec:expert-validation-groundedness}. 
We randomly sample 10 admissions to annotate, with 5 admissions coming from the \textit{grounded} test set overlapping with \verifactbhc{} and the remaining 5 admissions coming from the \dataset{} test set. 

For each admission, we ask each annotator to compare the BHCs generated after SFT on the (i) raw post-edit BHC, (ii) \textit{Grounded} filtered post-edit BHC, and (iii) \textit{Grounded + Global} filtered post-edit BHC against the zero-shot-generated BHC, providing one of the following ratings:
\begin{itemize}[topsep=0.1ex,itemsep=0.1ex]
    \item \textbf{A > B:} Clinical details in draft A are better grounded in the prior EHR notes.
    \item \textbf{A < B:} Clinical details in draft B are better grounded in the prior EHR notes.
    \item \textbf{A $\approx$ B:} The two drafts exhibit similar levels of groundedness. They are either \textit{similarly grounded} or \textit{similarly ungrounded} in the prior EHR notes, but one draft is not significantly worse than the other in terms of groundedness. 
\end{itemize}
A and B refer to either the zero-shot-generated BHC or one of the three BHCs generated after SFT on raw or filtered post-edit BHCs, which are provided to the annotators in fully randomized order. 

Below, we include the full annotation guide provided to the annotators:
\begin{promptbox}[Annotation Guide]
We are assessing the groundedness / faithfulness of LLM-drafted discharge summaries (specifically its brief hospital course (BHC) section) to the input context provided to the LLM. The BHC section details key clinical events, diagnoses, and treatments for a patient during their hospitalization. Since an LLM drafts the BHC based on all of the EHR notes available for a patient until their discharge from the hospital, a BHC is considered grounded if the clinical details present are well-substantiated by the prior EHR notes.\\

Given a pair of BHCs drafted by two different LLMs for the same hospital admission, the goal of this annotation task is to determine the model that produces better grounded BHCs. \\

\textcolor{blue}{\textbf{What You Will See}}\\
Each annotation example will include the following items:\\
- Two drafts of the BHC section for the same hospital admission\\
- Prior EHR notes available until discharge time\\

Consecutive annotation examples may come from the same hospital admission, in which case they share the same prior notes to assess groundedness against.\\

\textcolor{blue}{\textbf{What To Do}} \\
For each pair of BHCs, provide the following information:\\

1. “Groundedness” Rating\\
- A > B: Clinical details in draft A are better grounded in the prior EHR notes.\\
- A < B: Clinical details in draft B are better grounded in the prior EHR notes.\\
- A $\approx$ B: The two drafts exhibit similar levels of groundedness. They are either similarly grounded or similarly ungrounded in the prior EHR notes, but one draft is not significantly worse than the other in terms of groundedness.\\

2. Free-Text Comment\\
Please briefly explain your reasoning.\\

\textcolor{blue}{\textbf{Important Notes (Please Read)}}\\
- Compare the two drafts only based on their levels of groundedness, and not on whether one draft is e.g., more clinically useful, thorough, better organized than the other.
If two drafts exhibit similar levels of groundedness, do not account for the clinical significance, severity, or magnitude of deviations from the prior EHR notes to break the tie. For instance, if both drafts each respectively have one atomic clinical detail that is unsubstantiated, then they should likely be labeled as “A $\approx$ B”. \\
- Because all BHCs have undergone a de-identification process, the BHC drafts may contain redactions (e.g., “\_\_\_\_\_”, “**First Name4**”, “**Hospital1**”, “NamePattern2”) or imputed dates (e.g., “[**2122-6-20**]”) that render them awkward or grammatically incorrect. Please ignore such de-identification artifacts when comparing the drafts.

\end{promptbox}

\subsection{Inter-Annotator Agreement}
\label{appendix:groundedness-validation-agreement}

\begin{table}[!h]
  \centering
  \small
  \caption{Nurse-practitioner (NP) groundedness preferences over 10 admissions, where the BHCs from each SFT setting (\textbf{Step 2}, \textit{Grounded}, \textit{Grounded + Global}) are compared against the zero-shot-generated BHCs from the base model.
  Labels are consensus verdicts over three independent reads; comparisons on which
  the three reads disagreed entirely are counted as ties.}
  \begin{tabular}{lccc}
  \toprule
  \multirow{2}{*}{SFT Target} & \multicolumn{3}{c}{Groundedness vs.\ Base Model} \\
  \cmidrule(lr){2-4}
   & Better & Equal & Worse \\
  \midrule
  \textbf{Step 2} & 20\% (2/10) & 0\% (0/10) & 80\% (8/10) \\
  \midrule
  Filtered: \textit{Grounded} & 30\% (3/10) & 60\% (6/10) & 10\% (1/10) \\
  Filtered: \textit{Grounded + Global} & 50\% (5/10) & 10\% (1/10) & 40\% (4/10) \\
  \midrule
  All comparisons & 33.3\% (10/30) & 23.3\% (7/30) & 43.3\% (13/30) \\
  \bottomrule
  \end{tabular}
  \label{tab:task2-groundedness}
\end{table}

As discussed in Section~\ref{sec:expert-validation-groundedness}, for every comparison of groundedness between a BHC generated by one of the SFT models and the BHC zero-shot-generated by the base model, we obtain three independent NP annotations. 
Across $10 \times 3 = 30$ total pairwise comparisons (10 admissions, 3 comparisons per admission), we observe unanimous agreement on 63\% (19/30) of cases, majority agreement on 27\% (8/30), and a three-way disagreement in the remaining 10\% (3/30) of cases.
Since a three-way disagreement indicates that all three possible ratings---A > B, A $\approx$ B, A < B---are given, we treat the consensus label in such cases as a tie (i.e., A $\approx$ B). In Table~\ref{tab:task2-groundedness}, we also provide a detailed breakdown of the comparison results. 

\renewcommand\thetable{G\arabic{table}}
\renewcommand\thefigure{G\arabic{figure}}
\setcounter{table}{0}
\setcounter{figure}{0}
\section{Additional Results}
\label{appendix:results}
\subsection{SFT on Clinician-Written BHCs and Step 1 BHCs}
\label{appendix:sft-clinician-step1}

\begin{figure*}[!t]
    \centering
    \begin{tabular}{@{}c@{}}
        \begin{subfigure}{\linewidth}
            \includegraphics[width=\linewidth]{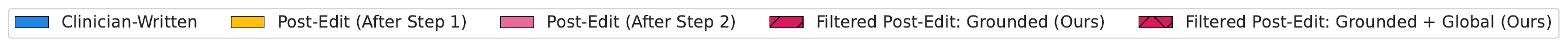}
        \end{subfigure} \\
        \begin{subfigure}{\linewidth}
            \includegraphics[width=\linewidth]{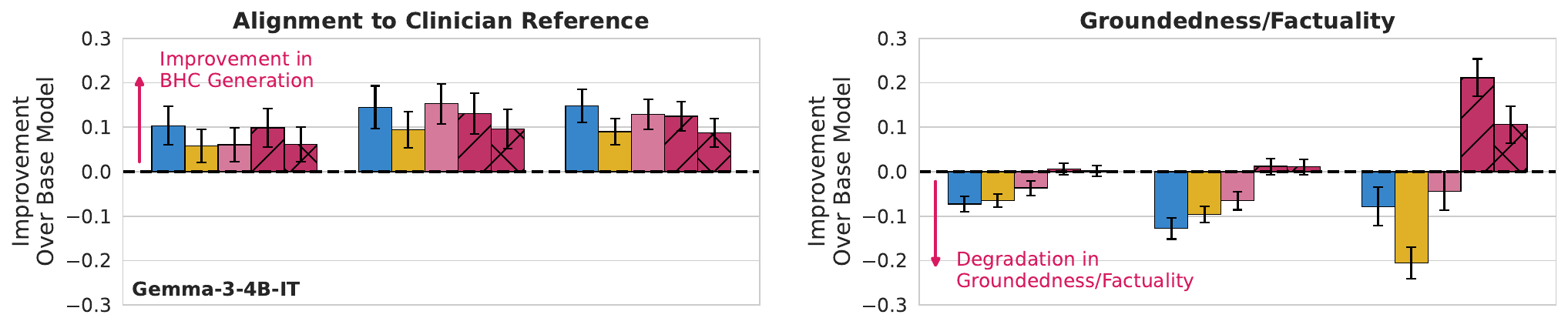}
        \end{subfigure} \\
        \begin{subfigure}{\linewidth}
            \includegraphics[width=\linewidth]{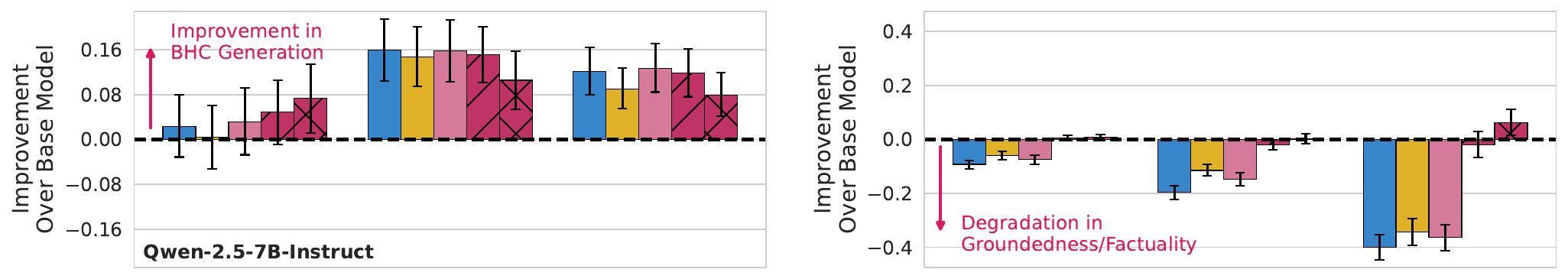}
        \end{subfigure} \\
        \begin{subfigure}{\linewidth}
            \includegraphics[width=\linewidth]{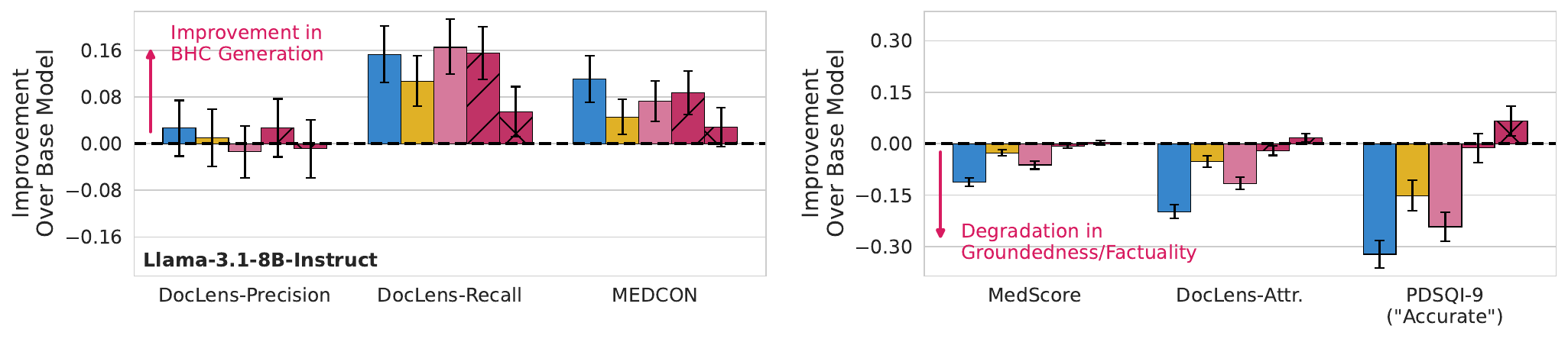}
        \end{subfigure}
    \end{tabular}
    \vspace{-10pt}
    \caption{SFT on the post-edit BHCs filtered using our edit tagger (Section~\ref{sec:filtering}) enables performance improvements (left) while preserving groundedness/factuality with respect to the input prior EHR notes (right). \textit{Clinician-Written} denotes the setting when we fine-tune each model on the original clinician-written BHCs for each admission. \textit{Post-Edit (After Step 1/2)} denotes the setting where we fine-tune on the BHCs resulting from \textbf{Step 1} and \textbf{Step 2} of our post-edit generation pipeline (Section~\ref{sec:dataset}). \textit{Filtered Post-Edit: Grounded} denotes the setting where we fine-tune on the post-edit BHCs from \textbf{Step 2}, after filtering out the ungrounded edits. \textit{Filtered Post-Edit: Grounded + Global} denotes the setting where we fine-tune on the post-edit BHCs from \textbf{Step 2}, after filtering out the ungrounded and personal edits.
    Error bars indicate the 95\% bootstrap CIs, computed by resampling the test set with replacement 10,000 times.}
    \label{fig:main-scores}
\end{figure*}

First, we consider fine-tuning on the original \textit{clinician-written} reference BHCs. 
This setting presumes access to human-written notes, which does not align with the main setting that we focus on, where clinicians always edit LLM drafts to complete documentation tasks (Section~\ref{sec:experiments}). 
We nonetheless include it as a reference point, as it reflects the gains achievable when fully human-written notes are available for supervision targets. 
As expected, we see that fine-tuning on clinician-written BHCs produces meaningful gains on alignment-based metrics across all 9 (model, metric) pairs relative to the base model (Figure~\ref{fig:main-scores}). 
Critically, we observe that these gains come at the cost of degraded groundedness across all 9 (metric, model) pairs. 
Such degradation may stem from a mechanism we note in Sections~\ref{sec:intro} and \ref{sec:taxonomy}: when clinicians draft or edit notes, they often draw information from sources external to the model's inputs, as noted in e.g., \citet{verifact}. 

Second, we assess the effect of using the intermediate (\textbf{Step 1}) output of our edit generation pipeline as the supervision target. In Figure \ref{fig:main-scores}, the trends mirror those of SFT on clinician-written BHCs and SFT on post-edit (\textbf{Step 2}) outputs: meaningful gains on alignment-based metrics across all models, but at the cost of degradation on all groundedness-based metrics. 
Relative to \textbf{Step 2}, the alignment gains are smaller for all 9 (metric, model) pairs. For groundedness, we observe all metrics degrade more than SFT on \textbf{Step 2} for Gemma-3-4B-IT, while the other two models show the opposite trend.

\subsection{Groundedness of Post-Edit BHCs in \dataset{}}
\label{appendix:train-groundedness}

\begin{figure*}[t!]
    \centering
    \includegraphics[width=0.95\linewidth]{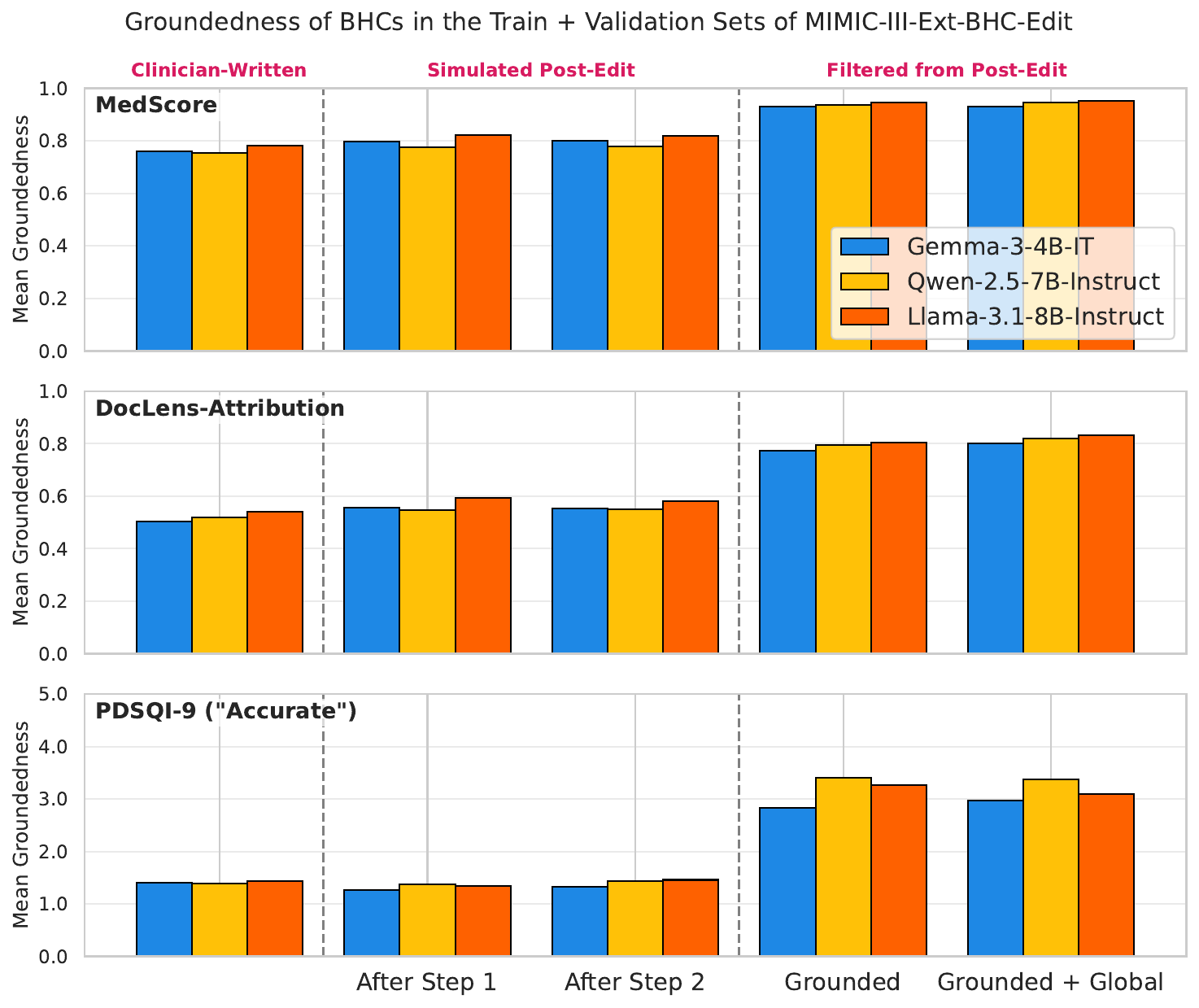}
    \vspace{-5pt}
    \caption{Differences in the average groundedness scores between the ground-truth BHCs used for SFT, when considering the clinician-written reference BHC (left), the simulated post-edit BHCs from \textbf{Step 1} and \textbf{Step 2} of the \dataset{} dataset construction process (middle), and the BHCs obtained from applying our \textit{Grounded} and \textit{Grounded + Global} edit filters (Section~\ref{sec:filtering}) (right). 
    We use the MedScore, DocLens-Attribution, and PDSQI-9 (``Accurate'') metrics considered in our experiments (Section~\ref{sec:experiments}), and show the scores on edit pairs generated for each base model. 
    The reference BHCs from MIMIC-III generally exhibit low groundedness scores, as suggested by prior works \citep{verifact}. 
    The simulated post-edit BHCs, which are aligned in content and style to the reference BHCs, also show low groundedness scores. 
    Filtering out the ungrounded edits leads to noticeably higher groundedness scores across all models and our two filtering strategies. 
    }
    \label{fig:train-groundedness}
\end{figure*}

In Figure \ref{fig:train-groundedness}, we analyze the differences in average groundedness scores between the ground-truth BHCs used for SFT, when considering the clinician-written BHC, the simulated post-edit BHCs from \textbf{Step 1} and \textbf{Step 2} of the \dataset{} dataset construction process, and the BHCs obtained from applying our \textit{Grounded} and \textit{Grounded + Global} edit filters (Section~\ref{sec:filtering}). 
We use the MedScore, DocLens-Attribution, and PDSQI-9 (``Accurate'') metrics considered in our experiments (Section~\ref{sec:experiments}), and show the scores on edit pairs generated for each base model. 
The clinician-written BHCs from MIMIC-III generally show low groundedness scores, as suggested by prior works \citep{verifact}. 
The simulated post-edit BHCs, which are aligned in content and style to the clinician-written BHCs, also exhibit low groundedness. 
Filtering out the ungrounded edits leads to noticeably higher groundedness scores across all models and our two filtering strategies. 

\subsection{Varying the Level of Ungroundedness in Training Targets}
\label{appendix:dose-response}

Our main experiments in Section~\ref{sec:results} compare between two fine-tuning setups: 
(i) SFT on raw post-edit BHCs, which contain all ungrounded edits simulated via our dataset construction process (Section~\ref{sec:dataset}); 
and (ii) SFT on the filtered post-edit BHCs (\textit{Grounded}, \textit{Grounded + Global}), where all edits classified as ungrounded by the edit tagger have been reverted (Section~\ref{sec:edit-tagger}). 
To characterize how the level of ungroundedness in the training targets affects SFT performance between these two extremes, we conduct additional experiments, where we revert a randomly selected fraction of the ungrounded edits identified by our edit tagger in each post-edit BHC, at fixed levels of 0\%, 30\%, 50\%, 75\%, and 100\%, using the same filtering procedure as in Appendix~\ref{appendix:edit-filtering-details}. 
The 0\% setting corresponds to SFT on the raw post-edit BHCs (\textbf{Step 2}), and the 100\% setting corresponds to applying the \textit{Grounded} filter as in Section~\ref{sec:edit-tagger}. 
We conduct this analysis with Qwen-2.5-7B-Instruct, using the same fine-tuning and evaluation setup as in Section~\ref{sec:experiments}.

\begin{table}[!h]
\centering
\small
\caption{SFT results for Qwen-2.5-7B-Instruct when reverting increasing fractions of randomly subsampled ungrounded edits in the post-edit BHCs. 0\% corresponds to SFT on the raw post-edit BHCs (\textbf{Step 2}); 100\% corresponds to applying the \textit{Grounded} filter.}
\label{tab:dose-response}
\resizebox{\linewidth}{!}{
\begin{tabular}{lcccccc}
\toprule
& \multicolumn{3}{c}{Alignment to Clinician-Written Reference} & \multicolumn{3}{c}{Groundedness} \\
\cmidrule(lr){2-4} \cmidrule(lr){5-7}
Setting & DocLens-P & DocLens-R & MEDCON & MedScore & DocLens-Attr. & PDSQI-9 \\
\midrule
Base Model & 0.271 & 0.174 & 0.253 & 0.948 & 0.900 & 0.599 \\
\midrule
SFT (0\% removed; \textbf{Step 2}) & 0.303 & 0.331 & 0.380 & 0.874 & 0.753 & 0.236 \\
SFT (30\% removed) & 0.288 & 0.347 & 0.382 & 0.877 & 0.783 & 0.303 \\
SFT (50\% removed) & 0.314 & 0.340 & 0.372 & 0.905 & 0.811 & 0.349 \\
SFT (75\% removed) & 0.321 & 0.326 & 0.376 & 0.927 & 0.828 & 0.429 \\
SFT (100\% removed; \textit{Grounded}) & 0.320 & 0.325 & 0.371 & 0.953 & 0.880 & 0.580 \\
\bottomrule
\end{tabular}
}
\end{table}

Table~\ref{tab:dose-response} shows that all three groundedness metrics improve monotonically as more ungrounded edits are removed. 
At 100\% removal, the MedScore and PDSQI-9 scores recover to near or above the level of those achieved by the base model. 
In contrast, alignment to clinician-written references remains largely stable across all levels (e.g., MEDCON score between 0.371 and 0.382), 
and remains substantially above the base model throughout all fractions. 
These results suggest that ungrounded edits account for most of the groundedness degradation observed after SFT on raw post-edit BHCs (Section~\ref{sec:result-naive}), while contributing little to improvements in alignment to clinician-written references.

These results also indicate how errors made by the edit tagger, if present, can affect filtering. 
Failing to identify a subset of ungrounded edits can be considered equivalent to removing only a fraction of ungrounded edits, 
and the gradual improvement in groundedness suggests that filtering remains beneficial under moderate error rates in the edit tagging outputs, while improvements in the edit tagger's accuracy translate to additional gains. 

\subsection{Average Lengths of Model-Generated BHCs}
\label{appendix:avg-bhc-length}

\begin{table*}[!h]
      \centering
      \caption{Average BHC length (word counts) by base model and SFT setting, on all examples from the \textit{grounded} \verifactbhc{} test set and the \dataset{} test set.}
      \label{tab:bhc-len-all}
      \resizebox{\linewidth}{!}{%
      \begin{tabular}{llcccccccc}
      \toprule
      \multirow{2}{*}{Model} & \multirow{2}{*}{Setting} & \multicolumn{2}{c}{Mean / Std} & \multicolumn{2}{c}{Min / Max} & \multicolumn{3}{c}{Quartiles} \\
      \cmidrule(lr){3-4} \cmidrule(lr){5-6} \cmidrule(lr){7-9}
       & & Mean & Std & Min & Max & Q1 & Q2 & Q3 \\
      \midrule
      \multirow{5}{*}{Gemma-3-4B-IT} & Base Model & 192.9 & 69.5 & 29 & 851 & 153 & 182 & 221 \\
      \cmidrule(l){2-9}
      & SFT (Clinician-Written BHC) & 329.2 & 399.0 & 38 & 3751 & 132 & 234 & 380 \\
      \cmidrule(l){2-9}
       & SFT (Post-Edit) & 350.7 & 413.0 & 40 & 3031 & 145 & 259 & 393 \\
       & SFT (Filtered Post-Edit: \textit{Grounded}) & 295.3 & 386.3 & 45 & 3270 & 126 & 213 & 331 \\
       & SFT (Filtered Post-Edit: \textit{Grounded + Global}) & 310.7 & 379.4 & 76 & 3566 & 156 & 214 & 346 \\
      \midrule
      \multirow{5}{*}{Qwen-2.5-7B-Instruct} & Base Model & 109.3 & 41.0 & 40 & 389 & 81 & 100 & 127 \\
      \cmidrule(l){2-9}
      & SFT (Clinician-Written BHC) & 294.5 & 285.1 & 51 & 3013 & 135 & 245 & 397 \\
      \cmidrule(l){2-9}
       & SFT (Post-Edit) & 279.8 & 271.1 & 62 & 2611 & 138 & 223 & 347 \\
       & SFT (Filtered Post-Edit: \textit{Grounded}) & 219.9 & 139.9 & 50 & 800 & 103 & 178 & 320 \\
       & SFT (Filtered Post-Edit: \textit{Grounded + Global}) & 177.3 & 221.9 & 50 & 2887 & 90 & 123 & 195 \\
      \midrule
      \multirow{5}{*}{Llama-3.1-8B-Instruct} & Base Model & 257.2 & 148.9 & 81 & 2098 & 169 & 232 & 319 \\
      \cmidrule(l){2-9}
      & SFT (Clinician-Written BHC) & 373.7 & 437.0 & 37 & 2969 & 151 & 287 & 426 \\
      \cmidrule(l){2-9}
       & SFT (Post-Edit) & 499.3 & 542.9 & 91 & 3483 & 242 & 374 & 509 \\
       & SFT (Filtered Post-Edit: \textit{Grounded}) & 349.4 & 346.4 & 66 & 3327 & 176 & 310 & 413 \\
       & SFT (Filtered Post-Edit: \textit{Grounded + Global}) & 287.9 & 255.2 & 66 & 2989 & 167 & 232 & 349 \\
      \bottomrule
      \end{tabular}
      }
\end{table*}

In Table~\ref{tab:bhc-len-all}, we summarize the average lengths of BHCs generated by the base model and the different fine-tuned models, including the ``oracle'' setting where models are fine-tuned on the original clinician-written reference BHCs. Across all model families, we overall see that the BHCs generated from the fine-tuned models tend to be much longer in length than the BHCs zero-shot-generated from the base models, with the filtered settings resulting in slightly shorter BHCs compared to the raw post-edit and clinician-written reference SFT settings.

\subsection{Reference-Based NLG Metric Evaluations}
\label{appendix:nlg-eval}

\begin{table*}[h!]
\centering
\small
\caption{Average BERTScore, ROUGE-L, and BLEU-2 by base model and SFT setting, on all examples from the \textit{grounded} test set based on \verifactbhc{} (Section~\ref{sec:experiments}).}
\resizebox{\linewidth}{!}{%
\begin{tabular}{llccc}
\toprule
\textbf{Model} & \textbf{Setting} & \textbf{BERTScore} & \textbf{ROUGE-L} & \textbf{BLEU-2} \\
\midrule
\multirow{6}{*}{Gemma-3-4B-IT}
 & Base Model                          & 0.827 & 0.152 & 0.138 \\
 \cmidrule(l){2-5}
 & SFT (Clinician-Written BHC)                    & 0.855 & 0.279 & 0.257 \\
 \cmidrule(l){2-5}
 & SFT (Post-Edit: \textbf{Step 1})                        & 0.842 & 0.218 & 0.202 \\
 & SFT (Post-Edit: \textbf{Step 2})                      & 0.850 & 0.248 & 0.222 \\
 \cmidrule(l){2-5}
 & SFT (Filtered Post-Edit: \textit{Grounded})       & 0.851 & 0.257 & 0.256 \\
 & SFT (Filtered Post-Edit: \textit{Grounded + Global})  & 0.843 & 0.217 & 0.233 \\
\midrule
\multirow{6}{*}{Qwen-2.5-7B-Instruct}
 & Base Model                          & 0.836 & 0.173 & 0.094 \\
 \cmidrule(l){2-5}
 & SFT (Clinician-Written BHC)                       & 0.857 & 0.272 & 0.259 \\
 \cmidrule(l){2-5}
 & SFT (Post-Edit: \textbf{Step 1})                    & 0.845 & 0.234 & 0.201 \\
 & SFT (Post-Edit: \textbf{Step 2})                        & 0.856 & 0.274 & 0.247 \\
 \cmidrule(l){2-5}
 & SFT (Filtered Post-Edit: \textit{Grounded})        & 0.855 & 0.270 & 0.304 \\
 & SFT (Filtered Post-Edit: \textit{Grounded + Global})& 0.849 & 0.238 & 0.218 \\
\midrule
\multirow{6}{*}{Llama-3.1-8B-Instruct}
 & Base Model                         & 0.837 & 0.191 & 0.186 \\
 \cmidrule(l){2-5}
 & SFT (Clinician-Written BHC)                       & 0.858 & 0.280 & 0.271 \\
 \cmidrule(l){2-5}
 & SFT (Post-Edit: \textbf{Step 1})                     & 0.843 & 0.217 & 0.207 \\
 & SFT (Post-Edit: \textbf{Step 2})                      & 0.849 & 0.233 & 0.167 \\
 \cmidrule(l){2-5}
 & SFT (Filtered Post-Edit: \textit{Grounded})         & 0.855 & 0.270 & 0.248 \\
 & SFT (Filtered Post-Edit: \textit{Grounded + Global})& 0.843 & 0.218 & 0.234 \\
\bottomrule
\end{tabular}
}
\label{tab:nlg-metrics}
\end{table*}

In Table \ref{tab:nlg-metrics}, we report the performance using standard NLG metrics---BERTScore, ROUGE-L, BLEU-2---for all models and data configurations on all examples from the \textit{grounded} test set based on \verifactbhc{}. 
For all metrics, we use the implementations in the \texttt{evaluate} package from HuggingFace.
SFT on clinician-written references yields the highest BERTScore for all three models and the highest or near-highest ROUGE-L. Among edit-based settings, the \textit{Grounded}-filter setting performs best or comparably, and achieves the highest BLEU-2 overall for Qwen-2.5-7B-Instruct. 
Filtering ungrounded edits consistently lifts BLEU-2 and ROUGE-L over the base model with comparable performance in terms of BERTScore. 

\subsection{LLM Ablations for Edit Tagging and Filtering}
\label{appendix:ablation-tagging-filtering}

While we use a single LLM (GPT-5.5) for edit simulation (Section~\ref{sec:dataset}), tagging, and filtering (Section~\ref{sec:edit-tagger}) in our experiments, we conduct additional experiments to demonstrate that our findings in Section~\ref{sec:results} generally hold in settings where we use an alternative LLM for edit tagging and filtering. 
To do so, we replace GPT-5.5 with Claude Opus 4.8~\citep{opus-4-8} for both edit tagging and filtering, while still starting from the same GPT-5.5-simulated edit pairs in \dataset{} and using the same fine-tuning and evaluation setup as in Section~\ref{sec:experiments}. 
We then measure their agreement at the edit-tagging level and in terms of the SFT performance trends (i.e., filtered vs.~raw post-edit training) resulting from using each model for edit tagging and filtering. 
We set up the Opus 4.8 implementation to be compliant with the PhysioNet Credentialed Health Data License.

\begin{table}[!h]
\centering
\small
\caption{Agreement between edit taggers based on GPT-5.5 and Opus 4.8. Span similarity refers to the mean similarity (Jaccard score) of matched atomic edit spans; filter agreement is the fraction of matched edit pairs on which both taggers make the same keep/revert decision under each filter.}
\label{tab:opus_agreement}
\begin{tabular}{lccc}
\toprule
Base Model & Mean Span Similarity & \textit{Grounded} & \textit{Grounded + Global} \\
\midrule
Gemma-3-4B-IT         & 0.909 & 0.820 & 0.702 \\
Qwen-2.5-7B-Instruct  & 0.910 & 0.820 & 0.701 \\
Llama-3.1-8B-Instruct & 0.904 & 0.832 & 0.711 \\
\bottomrule
\end{tabular}
\end{table}

\begin{table*}[!h]
\centering
\small
\caption{SFT results with GPT-5.5 vs.~Opus 4.8 as the edit tagger + filter, for Gemma-3-4B-IT and Qwen-2.5-7B-Instruct. Clinical claim/term-matching metrics (quantifying alignment to clinician-written references) are computed on the \textit{grounded} test set, while groundedness metrics are computed on examples from both test sets (Section~\ref{sec:experiments}). PDSQI-9 denotes the rescaled ``accurate'' score.}
\label{tab:opus_sft}
\resizebox{\textwidth}{!}{%
\begin{tabular}{llcccccc}
\toprule
& & \multicolumn{3}{c}{Alignment} & \multicolumn{3}{c}{Groundedness} \\
\cmidrule(lr){3-5} \cmidrule(lr){6-8}
Model & Setting & DocLens-P & DocLens-R & MEDCON & MedScore & DocLens-Attr. & PDSQI-9 \\
\midrule
\multirow{4}{*}{Gemma-3-4B-IT}
 & Base Model                  & 0.200 & 0.173 & 0.230 & 0.929 & 0.864 & 0.330 \\
 & SFT (\textbf{Step 2})             & 0.261 & 0.326 & 0.359 & 0.892 & 0.799 & 0.286 \\
 & SFT (\textit{Grounded}, GPT-5.5)  & 0.299 & 0.303 & 0.355 & 0.934 & 0.876 & 0.541 \\
 & SFT (\textit{Grounded}, Opus 4.8) & 0.274 & 0.329 & 0.357 & 0.918 & 0.851 & 0.420 \\
\midrule
\multirow{4}{*}{Qwen-2.5-7B-Instruct}
 & Base Model                  & 0.271 & 0.174 & 0.253 & 0.948 & 0.900 & 0.599 \\
 & SFT (\textbf{Step 2})             & 0.303 & 0.331 & 0.380 & 0.874 & 0.753 & 0.236 \\
 & SFT (\textit{Grounded}, GPT-5.5)  & 0.320 & 0.325 & 0.371 & 0.953 & 0.880 & 0.580 \\
 & SFT (\textit{Grounded}, Opus 4.8) & 0.322 & 0.299 & 0.373 & 0.932 & 0.835 & 0.463 \\
\bottomrule
\end{tabular}
}
\end{table*}

Table~\ref{tab:opus_agreement} shows that the two models largely agree on edit spans and on which edits each filter reverts. 
The atomic edit outputs from each model were matched by computing the Jaccard similarity score between text spans for every possible pair of atomic edits with the same edit operation (e.g., addition, deletion, substitution) and identifying the highest-scoring pairs. 
The ``Mean Span Similarity'' in Table~\ref{tab:opus_agreement} refers to the average of these Jaccard similarity scores across all matched pairs. 
Their agreement in edit tagger outputs is generally lower with the \textit{Grounded + Global} filter, consistent with the lower expert agreement on the \textit{Global vs.~Personal} axis (Section~\ref{sec:edit-tagger}). 
With either model, SFT on \textit{Grounded}-filtered BHCs substantially reduces the loss in groundedness caused by SFT on raw post-edit BHCs, while retaining most of the alignment gains (Table~\ref{tab:opus_sft}). 
However, the Opus 4.8-based filter recovers groundedness less fully: its scores remain slightly below the base model on MedScore and DocLens-Attribution for both models, and on PDSQI-9 for Qwen-2.5-7B-Instruct. 
Nonetheless, our overall conclusions in Section~\ref{sec:results} remain unchanged: \textit{with both GPT-5.5 and Opus 4.8-based edit taggers, filtering out the ungrounded edits mitigates degradation in groundedness after SFT, while enabling improvements over the base model}. 

\end{document}